\documentclass[11pt]{article}

\usepackage[margin=1in]{geometry}
\usepackage{amsmath,amssymb}
\usepackage{booktabs}
\usepackage{multirow}
\usepackage{graphicx}
\usepackage{xcolor}
\usepackage[colorlinks=true,linkcolor=blue,citecolor=blue,urlcolor=blue]{hyperref}
\usepackage{natbib}

\usepackage[utf8]{inputenc}
\usepackage[T1]{fontenc}
\usepackage[english]{babel}

\title{Beyond Depth Truncation: Controlled Evaluation of Depth Utilization in Recursive Language Models}

\author{
  Ha Van Dau\textsuperscript{1} \qquad
  Thanh Tung Khuat\textsuperscript{2} \qquad
  Nguyen Thanh Dung\textsuperscript{3} \\[0.6em]
  \textsuperscript{1}Blaze AI, \texttt{vand@blaze.vn} \\
  \textsuperscript{2}NuverxAI - AI \& Creative Innovation Company Limited, \texttt{thanhtung.khuat@nuverxai.com} \\
  \textsuperscript{3}Ho Chi Minh City University of Technology (HCMUT), \texttt{thanhdungng04@gmail.com}
}
\date{\today}

\begin{document}
\maketitle

\begin{abstract}
Depth-recurrent language models iteratively apply a small layer stack, decoupling per-token compute from distinct parameter count. To determine whether such a model genuinely utilizes its depth, both recurrence and layer-pruning literatures rely on a shared evaluation: truncating depth at inference time, plotting quality against retained depth fraction, and reading off the slope. While cheap and training-free, this metric suffers from an unexamined flaw: it extracts a single scalar from an intervention that alters multiple model properties simultaneously. Depth truncation concurrently reduces the number of block applications, decreases the volume of distinct computation performed, and pushes the readout head onto an out-of-distribution residual stream. The observed slope conflates all three factors, yet is conventionally interpreted as reflecting solely the second.

We propose the Depth Control Protocol (DCP), a diagnostic suite that disentangles these three quantities. DCP comprises three positive controls that isolate each factor while varying the others, a negative control applying the identical interventions to dense transformers to ensure the effect is not an artifact of the measurement protocol, and a controlled training intervention to verify causality. The linchpin control, running the full budget of block applications while executing only a single distinct iteration, is strictly realizable only in depth-wise weight-sharing architectures, since in a dense network repeating a layer yields an entirely different model rather than the same model in an alternative configuration.
\end{abstract}

% =====================================================================
\section{Introduction}
\label{sec:intro}

Depth-recurrent language models iteratively apply the same small stack of layers to their own representations, rather than stacking numerous layers with unique parameters \citep{dehghani2018universal, geiping2026scaling}. Through depth-wise weight sharing, per-token computation is decoupled from the number of distinct parameters: the same set of weights can be unrolled for two iterations or thirty-two iterations. This property has been exploited along two primary applications. The first is inference-time compute scaling, spending additional iterations on challenging inputs without enlarging the model footprint. The second is latent-space reasoning, executing multiple transformation steps prior to token generation rather than externalizing reasoning traces into explicit text. Both paradigms are especially appealing at modest scale: if depth can partially substitute for data, a model trained on far fewer than trillions of tokens can still attain nontrivial reasoning capabilities.

Both directions necessitate answering the same empirical question: does a given model genuinely utilize its depth? At present, this question is answered via an evaluation shared across both recurrence and layer-pruning literature \citep{gromov2025unreasonable, men2025shortgpt}: truncating depth at inference, plotting quality as a function of retained depth fraction, and interpreting the slope as the contribution of depth. This metric is cheap and requires no retraining, which has cemented it as the default across both lines of work. However, the shared limitation of current evaluations is that they collapse a multi-factorial intervention into a single scalar, subsequently attributing that scalar entirely to a single attribute. This paper asks what this metric actually measures.

\begin{figure}[t]
\centering
\includegraphics[width=\linewidth]{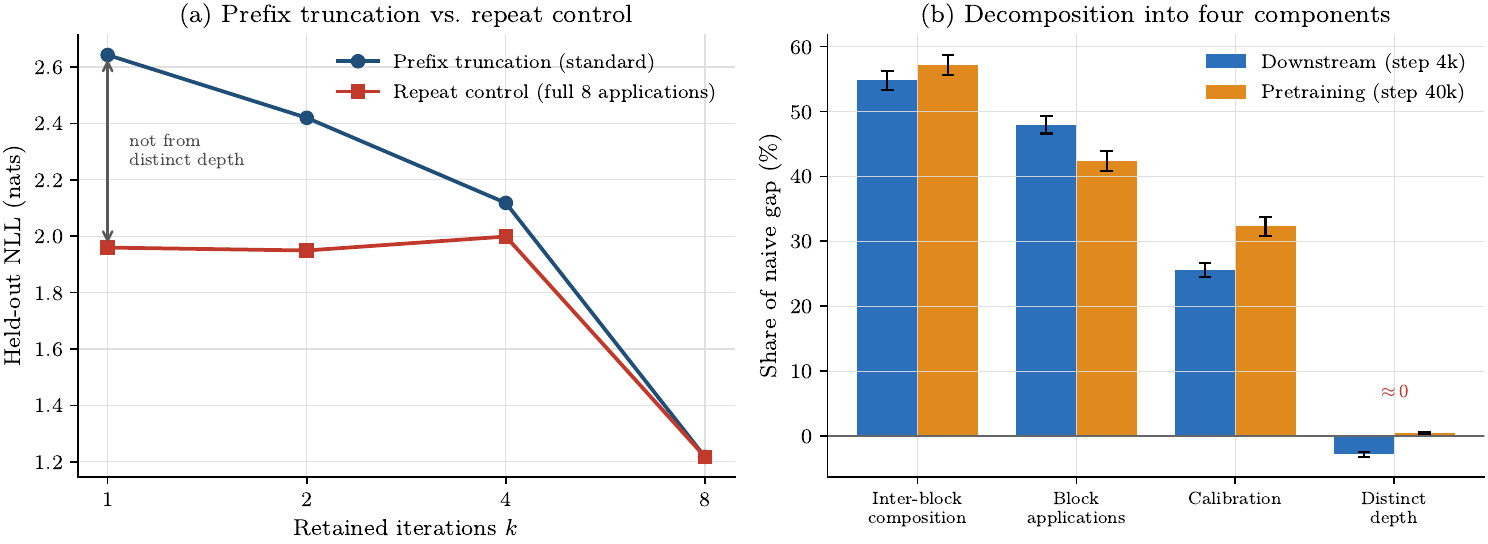}
\caption{Summary of main results. \textbf{(a)} Prefix truncation, the standard metric, yields a monotonic decrease in NLL as the number of retained iterations increases. The repeat control executes the full $8$ block applications at every $k$ while varying only the number of \emph{distinct} iterations; it remains nearly flat across $k = 1, 2, 4$. The gap between the two curves at $k = 1$ therefore does not arise from distinct depth. \textbf{(b)} Decomposition of the naive gap into four constituent components across two checkpoints; error bars denote $95\%$ bootstrap confidence intervals over $5{,}000$ document resamplings. The distinct depth component is approximately zero in both checkpoints. The suffix control is omitted from panel (a) for visual clarity; the complete three-control suite is presented in Figure~\ref{fig:controls}.}
\label{fig:teaser}
\end{figure}

The root cause of this limitation is that depth truncation conflates at least three distinct quantities. First is how many \emph{block applications} write into the residual stream. Second is how much \emph{distinct computation} these applications execute. Third is whether the readout head remains calibrated to the residual distribution it receives, as depth truncation shifts the readout into a residual distribution unseen during training. The observed slope reflects the sum of all three, yet in practice it is interpreted as though it isolates only the second.

\paragraph{Proposed Method.}
We propose the Depth Control Protocol (DCP), a diagnostic procedure that disentangles these three quantities and validates the underlying causal mechanisms. DCP consists of three core components. The first component is a suite of three \emph{positive controls}, each holding one quantity fixed while manipulating the others, thereby attributing each portion of the apparent gap to its rightful source. The linchpin control in this set runs the full budget of block applications while executing only a single distinct iteration; it is \emph{strictly realizable} only under depth-wise weight sharing, because in a dense network repeating the $k$-th layer eight times produces a different model rather than the same model under an alternative runtime configuration. The second component is a \emph{negative control} applying the exact same interventions to dense transformers, serving to distinguish genuine confounders from artifacts of the measurement protocol itself. The third component is a \emph{controlled training intervention} that perturbs the single suspected causal variable, thereby turning observed correlations into verified causal relationships. Formal specifications of all three components are detailed in Section~\ref{sec:method}.

Applying DCP across five configurations reveals that naive truncation substantially overestimates the contribution of recurrent depth (Figure~\ref{fig:teaser}). The overestimated quantity is the NLL reduction on a held-out set when unrolling additional iterations---the precise metric underpinning claims that ``depth buys performance''. On a 542.8M parameter recurrent model trained for mathematical reasoning, naive truncation attributes a reduction of $1.4251$~nats to latent depth. Our three positive controls reveal that $47.9\%$ of this gap stems from the number of block applications rather than the distinct computation they perform, with an additional $25.6\%$ attributable to readout miscalibration; additional distinct iterations within a reasoning block contribute $-2.8\%$ $[-3.2, -2.4]$, indicating that under this control they yield no NLL improvement.

Negative controls and training interventions demonstrate that this finding is not idiosyncratic to a single model. Applying the identical control suite to layer truncation on two standard transformers reverses the sign of the ``block application'' component, proving that these confounders are not measurement artifacts. Furthermore, when evaluated on Huginn-0125---a public recurrent model whose iterations were \emph{sampled during training} and which features published claims of inference-time compute scaling---the calibration confounder disappears entirely: its fitted temperature remains flat within $0.045$ across a $32\times$ depth range, compared to $0.57$ across an $8\times$ range in the fixed-depth model. Across four models, the confounder tracks exactly one variable: whether the readout head witnessed more than a single recurrent depth during training, independent of model family or parameter scale. The vulnerability identified by these results thus resides not in the recurrent architecture itself, but in the \emph{training schedule}: a model trained strictly at a single depth develops a readout specialized to that depth, causing any truncation evaluation to conflate specialization drift with the genuine contribution of depth.

These two components possess different scopes, which we explicitly delineate. The \emph{application count} confounder is unique to weight-sharing architectures: its sign inverts when applied to dense transformers. The \emph{calibration} confounder, by contrast, is not: evaluated on Qwen2.5-Math-1.5B with general text where the model is well calibrated ($T = 1.03$ at full depth), layer truncation still inflates the fitted temperature to $1.91$, accounting for $16.5\%$ of the apparent gap. Consequently, any study drawing conclusions from performance curves across retained layers without recalibration reports a metric contaminated by this component, even for non-recurrent architectures.

These findings yield two practical implications. Randomly sampling recurrent depth during training eliminates this confounder at zero inference cost. Until this practice becomes standard, depth truncation studies on models trained at fixed depth should report DCP alongside naive curves, as naive curves alone exaggerate the depth effect by roughly a factor of two.

\paragraph{Contributions. Our main contributions in this research include:}
\begin{enumerate}
  \item We propose the Depth Control Protocol (DCP), a diagnostic suite comprising three positive controls that decompose depth truncation into its constituent quantities, alongside a negative control and a training intervention (Section~\ref{sec:method}). We formally state its applicability conditions, prove the decomposition is exact and exhaustive, and outline five identification limits, including path dependency within the configuration space (Section~\ref{sec:conditions}).
  \item Using the three positive controls, we identify and quantify two primary confounders in depth truncation: block application count and calibration drift (Section~\ref{sec:depth}).
  \item We decompose the $1.4251$~nat naive depth effect into $47.9\%$ block applications, $-2.8\%$ distinct depth, and $54.8\%$ inter-block composition, with bootstrap confidence intervals excluding zero across all components. The finding that distinct depth contributes approximately zero is replicated across three checkpoints spanning distinct training regimes, including a reinforcement-learning checkpoint ($-0.14\%$, Section~\ref{sec:replication}).
  \item We introduce a negative control on dense transformers demonstrating that these confounders are intrinsic properties of the evaluated models rather than measurement artifacts: applying identical interventions to layer truncation reverses the sign of the ``block application'' component (Section~\ref{sec:negctrl}).
  \item We employ cross-model comparisons to validate the underlying generative mechanism, specifically readout specialization to a single training depth. On Huginn-0125, where recurrence depth is sampled during training, the calibration confounder vanishes ($-0.6\%$ vs.\ $25.6\%$); across four models, this confounder tracks this operational variable rather than architecture family or parameter scale (Section~\ref{sec:huginn}).
  \item We introduce a controlled training intervention establishing causal attribution: $1{,}000$ steps of continued training with sampled depth reduces the calibration share from $29.9\%$ to $5.1\%$, whereas a control branch trained for the identical budget at fixed depth remains virtually unchanged (Section~\ref{sec:sampledepth}).
  \item We characterize the dynamics of the recurrent stack across five configurations (Section~\ref{sec:dynamics}). Only Huginn-0125 functions as a contraction mapping, explaining its empirical resilience; the remaining four configurations, including both dense transformers, amplify perturbations at comparable rates. Sampled-depth training does not alter this dynamical property, demonstrating that calibration and operator dynamics are distinct mechanisms.
  \item We formulate a concrete remedial intervention---sampling recurrence depth during training across the target deployment configuration space---and recommend reporting DCP alongside any truncation-based depth claims. Our evaluation toolkit is made publicly available with this paper.
\end{enumerate}

% =====================================================================
\section{Related Work}
\label{sec:related}

\paragraph{Adaptive Computation and Recurrent Depth.} Adaptive Computation Time (ACT)
\citep{graves2016adaptive} introduced variable per-token compute via a learned halting
distribution; PonderNet \citep{banino2021pondernet} reformulated halting through a probabilistic objective.
Universal Transformers \citep{dehghani2018universal} combined depth-wise weight sharing
with ACT. Huginn-0125 \citep{geiping2026scaling} scaled a weight-tied recurrent core
to 3.5B parameters, sampling recurrence steps during training---a pivotal operational feature for this study.
Our evaluated architecture differs by chaining \emph{two} distinct reasoning blocks rather than repeating a single block,
which renders the distinction between ``distinct computation'' and ``repeated unrolling'' empirically measurable.

This research line has expanded rapidly in 2026. Recent proposals investigate
\emph{what} to repeat \citep{lin2026allocating}, residual normalization under tied weights \citep{li2026deeploop},
gated modulation to prevent representational collapse \citep{hegazy2026recurrentgpt}, and retrofitting recurrent depth
into pretrained dense models \citep{shapiro2026retrofitting}. Crucially, all of these works report model quality as a function
of iteration count. This family of performance curves is precisely what this paper investigates.

\paragraph{Dynamics of Recurrent Operators.} A parallel inquiry explores
\emph{when} unrolling additional iterations is advantageous, answering through the dynamical properties
of the trained operator. \citet{viakhirev2026think} categorized operators into settled, boundary, and drifting regimes,
establishing sufficient conditions under which added depth preserves solution fidelity. \citet{zhang2026does} investigated
when tied recurrence faithfully implements an algorithm, identifying a compute-budget law linking execution speed to the training contract.
SCORE \citep{godin2026score} went further by enforcing contractivity by construction via an ODE-style update.

We explicitly clarify our conceptual overlap with these works. Our contraction coefficient measurements in Section~\ref{sec:dynamics}
and the regime classifications of \citet{viakhirev2026think} probe the same underlying operator properties, arriving at concordant qualitative conclusions:
contractive operators exhibit resilience to depth variations, whereas amplifying operators do not. The divergence lies in the direction of the core inquiry:
they ask what happens when depth is \emph{added} beyond the training budget; we ask where the performance drop observed when depth is \emph{removed} actually originates.
The primary contribution of this work lies not in dynamical characterization, but in decomposing the apparent performance gap into constituent quantities---a separation that dynamical analysis alone cannot achieve.

\paragraph{Diagnostics for Recurrent Computation.} Closest in philosophical posture is
\citet{lam2026ignition}, which directly asked whether reported latent reasoning represents genuine computation or an \emph{instrument artifact},
utilizing pre-registered probe gates and simultaneous measurements across readout and latent channels. They concluded the phenomenon was genuine and localized at the readout.
\citet{lin2026allocating} introduced the Iteration Transfer Ratio (ITR) to quantify the non-redundant contribution of each iteration.

These metrics and DCP address fundamentally different questions, a distinction we emphasize as critical. ITR identifies \emph{which iterations warrant unrolling} for architectural design. \citet{lam2026ignition} interrogated whether a specific empirical phenomenon is real. DCP asks whether a \emph{widely adopted evaluation protocol} correctly attributes credit, answering by decomposing the apparent gap into interventional components. Notably, all three converge on the same locus: the readout. In their work, the readout is where genuine computation crystallizes; in ours, the readout is where calibration drift accumulates and is mistakenly counted as the contribution of depth.

\paragraph{Latent Reasoning.} Quiet-STaR \citep{zelikman2024quiet} trains
latent rationales via REINFORCE objectives; Coconut \citep{hao2024training} conducts multi-step reasoning continuous latent space rather than language token space. These paradigms share the goal of non-verbalized internal computation with recurrent models, but vary the number of latent \emph{tokens} rather than the number of \emph{block applications} of a layer stack; hence, our identified confounders do not manifest in the same form.

\paragraph{Layer Pruning and Early Exit.} Depth truncation is standard across layer pruning,
where the metric of interest is residual performance after layer ablation \citep{gromov2025unreasonable, men2025shortgpt}, and across early-exit models,
where inference halts at intermediate layers conditioned on inputs \citep{elbayad2019depth, schuster2022confident}. Both paradigms manipulate \emph{which} and \emph{how many} layers execute.
This literature is likewise undergoing self-scrutiny: \citet{wang2025fewer} showed that pruning just one or two layers shatters inference-time compute scaling, while \citet{shi2026understanding} traced catastrophic collapse to sharp transitions in decision representations. Both draw conclusions from performance curves over remaining layers.

The fundamental divergence from our work lies here. In a dense network, each layer is a unique function; hence, no configuration of ``$L$ applications, one distinct computation'' exists: repeating layer $k$ creates a different model rather than the same model under an alternate configuration. Our repeat control (Section~\ref{sec:controls}) requires depth-wise weight sharing and cannot be constructed within the setting of those works. This is why our protocol isolates application count from distinct computation, a separation that dense layer pruning cannot achieve in principle.

\paragraph{Training with Stochastic Depth.} Stochastic depth
\citep{huang2016deep} and LayerDrop \citep{fan2019reducing} randomly drop layers during training, with LayerDrop explicitly targeting training-free inference pruning. The intervention we explore in Section~\ref{sec:sampledepth}---sampling recurrence depth during training---belongs to this conceptual family.

We make no claim of novelty regarding this training technique. The contribution of this paper operates on a different plane: prior works propose a \emph{training technique} to induce prunability, whereas we interrogate whether an \emph{evaluation metric} is interpretable, demonstrating that fixed-depth training invalidates that metric in a quantifiable manner. Consequently, the two lines converge: the technique originally proposed to enhance prunability is precisely what restores the validity of the evaluation metric. An open cell in our design---evaluating our control suite on a dense model trained with LayerDrop---is discussed in Section~\ref{sec:controlled}.

\paragraph{Calibration.} Temperature scaling \citep{guo2017calibration} is the
standard single-parameter post-hoc calibration technique. We employ it to \emph{quantify} the magnitude of miscalibration, not to patch the model: the fitted temperature measures how far the output distribution departs from the distribution it was trained upon. Early-exit literature also examines calibration at intermediate exit points \citep{schuster2022confident}, but treats calibration as a \emph{control signal} to govern halting decisions. Here, calibration serves as a \emph{measurement probe}, and the specific profile we observe (fitted temperature remaining flat across truncated depths before dropping sharply to $\approx 1$ strictly at the training depth---a step function rather than a gradual drift) acts as an empirical diagnostic signature rather than generic miscalibration.

\paragraph{Position Embeddings and Evaluation Data.} The evaluated model utilizes rotary
position embeddings (RoPE) \citep{su2024roformer} with a dual-stream variant detailed in Section~\ref{sec:setup}. We evaluate on held-out mathematical text; benchmark contamination in this domain is an established concern \citep{brown2020language, tola23}, and our evaluation corpus was curated following a rigorous decontamination audit. All reported confidence intervals represent percentile bootstrap intervals \citep{tibshirani1993introduction}.

% =====================================================================
\section{Proposed Method: Depth Control Protocol (DCP)}
\label{sec:method}

This section details the Depth Control Protocol (DCP), our proposed diagnostic procedure. The problem DCP addresses is: given a quality-versus-depth curve obtained via depth truncation, disentangle the apparent gap into its constituent causal sources and determine which source genuinely reflects the contribution of depth. All three positive controls intervene exclusively at inference time over frozen weights; their computational cost is of the same order as the naive truncation they augment.

\paragraph{Notation.} Consider a recurrent model comprising $n_b$ reasoning blocks, each applied $n_i$ times, yielding a total budget of $N = n_b \cdot n_i$ block applications. Standard prefix truncation retains the first $k \le N$ applications and records performance $Q(k)$; the apparent gap is $\Delta_{\text{naive}} = Q(1) - Q(N)$. To ensure learned per-iteration embeddings remain aligned with training, the original global iteration index $g_{\text{idx}} = b \cdot n_i + t$ is preserved for block $b$ at iteration $t$, rather than re-indexing from zero.

\subsection{Component 1: Three Positive Controls}
\label{sec:controls}

The three interventions below each isolate one quantity while holding the remaining quantities constant, thereby attributing each portion of $\Delta_{\text{naive}}$ to its rightful source.

\paragraph{Repeat Control (repeat).} Execute the first $k$ distinct applications, then repeat the $k$-th application until completing $N$ total block applications. This control holds \emph{block applications} constant, preserving both the norm and the empirical distribution of the residual stream feeding into the readout head exactly as seen during training, while varying solely the volume of \emph{distinct} computation. This represents the linchpin control of DCP and is \emph{strictly realizable} only in depth-wise weight-sharing architectures: in a dense network, repeating layer $k$ multiple times constructs a different function rather than evaluating the same model under an alternate execution schedule.

\paragraph{Suffix Control (suffix).} Execute the \emph{final} $k$ applications rather than the initial ones. This control decouples ``how many iterations'' from ``which specific iterations'', thereby verifying whether iteration count is indeed the governing causal variable.

\paragraph{Temperature Calibration Control.} Refit a single logit temperature $T$ for each depth $k$ on a held-out calibration set and report post-calibration performance. Because the final layer normalization and readout head are trained on the residual statistics of full forward passes, truncation miscalibrates them for reasons unrelated to reasoning capacity; recalibration isolates this distributional drift from underlying computational capability. We fit $T$ via L-BFGS with strong-Wolfe line search on $\log T$ \citep{guo2017calibration}, resolving optimal values to $<10^{-3}$; a coarse grid search is inadequate here because temperature drifts compared across architectures differ by an order of magnitude. We emphasize that temperature scaling is used here to \emph{quantify} the extent of miscalibration, not to modify model predictions.

\subsection{Component 2: Decomposition of the Apparent Gap}
\label{sec:decomp}

Using the three controls above, $\Delta_{\text{naive}}$ is decomposed into four constituent components, each defined as the performance difference between two conditions differing in exactly one operational attribute:
\begin{align}
  \Delta_{\text{apply}}   &= Q_{\text{prefix}}(1) - Q_{\text{repeat}}(1),
    \label{eq:share-apply}\\
  \Delta_{\text{distinct}} &= Q_{\text{repeat}}(1) - Q_{\text{repeat}}(n_i),
    \label{eq:share-distinct}\\
  \Delta_{\text{compose}}   &= Q_{\text{repeat}}(n_i) - Q_{\text{repeat}}(N),
    \label{eq:share-compose}\\
  \Delta_{\text{calib}} &= \Delta_{\text{naive}}
    - \bigl(Q^{T}_{\text{prefix}}(1) - Q^{T}_{\text{prefix}}(N)\bigr),
    \label{eq:share-calib}
\end{align}
where $Q^{T}$ denotes performance after temperature recalibration at each $k$. The relative share of each component is defined as its ratio to $\Delta_{\text{naive}}$. The calibration component by definition overlaps with the first three components, as it is evaluated across the same boundary conditions ($k=1$ and $k=N$); we report it alongside the structural decomposition rather than summing into the total.

Because all components are differences across conditions and all shares are ratios of these differences, confidence intervals computed independently per condition would be misleading: conditions are evaluated on identical documents and are strongly correlated. DCP therefore mandates resampling document indices \emph{once per bootstrap iteration} and propagating that identical resample across all conditions and derived components \citep{tibshirani1993introduction}.

\subsection{Applicability Conditions and Identification Limits}
\label{sec:conditions}

This section explicitly formalizes the assumptions of DCP, identifying which quantities are identifiable and which are not. An empirical evaluation protocol is only sound when its failure modes are rigorously characterized.

\paragraph{Property 1 (Exact and Exhaustive Decomposition).}
The three components in~\eqref{eq:share-apply}--\eqref{eq:share-compose} form a telescoping sequence along the path
$\text{prefix}(1) \rightarrow \text{repeat}(1) \rightarrow \text{repeat}(n_i) \rightarrow \text{repeat}(N)$ within configuration space.
Because $Q_{\text{repeat}}(N) = Q_{\text{prefix}}(N)$ by definition (both executing the full forward pass), the sum of the three components collapses algebraically to
$Q_{\text{prefix}}(1) - Q_{\text{prefix}}(N) = \Delta_{\text{naive}}$.
The decomposition leaves no unassigned residual and requires no additivity assumptions regarding underlying effects.

\paragraph{Property 2 (Forced Path, Not Arbitrary Choice).}
A natural critique is that attribution shares depend upon the path traversed through configuration space, rendering individual labels arbitrary. This critique fails here: the geometry of realizable configurations forces exactly one path whose segments are strictly univariate.

Assigning coordinates $(a, d)$ to each configuration, where $a$ denotes block applications and $d$ denotes distinct iterations, the two truncation regimes map to two distinct loci:
\begin{align}
  \text{prefix}(k) &\;\longmapsto\; (a, d) = (k, k), \label{eq:coord-prefix}\\
  \text{repeat}(k) &\;\longmapsto\; (a, d) = (N, k). \label{eq:coord-repeat}
\end{align}
Prefix truncation \emph{couples} $a$ and $d$; repeat controls fix $a = N$ and vary only $d$. Consequently:
\begin{itemize}
  \item A step varying only $a$ requires two configurations with identical $d$ and different $a$. The unique valid pair is $\text{prefix}(k)$ and $\text{repeat}(k)$ at matching $k$.
  \item A step varying only $d$ requires two configurations with identical $a$ and different $d$. The unique valid pair is $\text{repeat}(k_1)$ and $\text{repeat}(k_2)$.
\end{itemize}
Any decomposition into strictly univariate steps can only utilize these two step types, and the path in~\eqref{eq:share-apply}--\eqref{eq:share-compose} is the unique sequence connecting $\text{prefix}(1)$ to $\text{repeat}(N)$. The path is not merely one among many valid choices; it is the sole valid choice.

\paragraph{Limit 1 (Confounded Paths Yield Meaningless Labels).}
This does not prevent one from constructing alternative paths; it simply means such paths must contain multivariate steps. The consequences are severe. Consider the path
$\text{prefix}(1) \rightarrow \text{prefix}(n_i) \rightarrow \text{repeat}(n_i) \rightarrow \text{repeat}(N)$: while its sum still telescopes to $\Delta_{\text{naive}}$, its first step transitions from $(1,1)$ to $(n_i, n_i)$, varying \emph{both} coordinates simultaneously. Attributing that step's difference entirely to distinct depth conflates $d$ with the simultaneous shift in $a$.

Table~\ref{tab:pathdep} demonstrates the numerical consequences on our calibration grid.

\begin{table}[t]
\centering
\small
\begin{tabular}{lrrrr}
\toprule
& \multicolumn{2}{c}{Univariate path (used in this paper)}
& \multicolumn{2}{c}{Confounded path} \\
\cmidrule(lr){2-3}\cmidrule(lr){4-5}
Branch & Application & Distinct & ``Application'' & ``Distinct'' \\
\midrule
A: fixed $8$              & $42.7\%$ & $-4.81\%$ & $15.2\%$  & $22.78\%$ \\
B: $\mathcal{U}\{1..8\}$    & $41.8\%$ & $+4.80\%$ & $-49.9\%$ & $96.47\%$ \\
C: $\mathcal{U}\{5..8\}$    & $72.8\%$ & $-1.60\%$ & $-17.5\%$ & $88.72\%$ \\
\bottomrule
\end{tabular}
\caption{Why univariate steps are mandatory. The right columns trace the path
$\text{prefix}(1) \rightarrow \text{prefix}(4) \rightarrow \text{repeat}(4) \rightarrow \text{repeat}(8)$, where the first step varies both block applications and distinct iterations simultaneously. While the sum remains exact, the ``distinct'' component absorbs nearly the entire gap, while the ``application'' component receives negative values in two branches, rendering the labels meaningless. Values evaluated at step $6{,}000$.}
\label{tab:pathdep}
\end{table}

Under the confounded path, the component labeled ``distinct depth'' surges from $-1.60\%$ to $88.72\%$ in branch C, while ``block applications'' turns negative in two of three branches. A decomposition where an application share drops below $-49\%$ does not represent an alternative interpretation; it signals that labels have lost their physical meaning. We highlight this because we initially constructed such a path during exploratory robustness checks; its $88.72\%$ figure appeared deceptively plausible before checking coordinate orthogonality.

\paragraph{Limit 2 (Composition Component is Inherently Mixed).}
The third step, $\text{repeat}(n_i) \rightarrow \text{repeat}(N)$, maintains $a = N$ while increasing $d$ from $n_i$ to $N$. In architectures with $n_b \ge 2$, this transition simultaneously engages the second reasoning block. This step is therefore \emph{not} strictly univariate in the narrow sense, which is why we designate it ``inter-block composition'' rather than distinct depth. Disentangling these two factors would require a configuration running $N$ applications with $d$ distinct iterations \emph{confined entirely within the first block}---an execution schedule not realizable under standard loop unrolling.

\paragraph{Condition C1 (Depth-Wise Weight Sharing).}
The repeat control assumes the same function $f_\theta$ is applied $N$ times, such that $h_{t+1} = f_\theta(h_t)$ with shared $\theta$. In dense networks, each layer is an independent function $f_{\theta_t}$; repeating layer $k$ constructs an \emph{entirely different model} rather than the same model under an alternate schedule. When C1 is violated, DCP is restricted to prefix, suffix, and calibration controls; this is why the negative control in Section~\ref{sec:negctrl} lacks the repeat component.

\paragraph{Condition C2 (Loop Index Preservation).}
If a model employs learned per-iteration embeddings, the repeat control must preserve the original global iteration index $g_{\text{idx}} = b \cdot n_i + t$. Re-indexing would introduce a second confounded variable, preventing attribution to distinct computation.

\paragraph{Condition C3 (Absence of Early Exit).}
DCP assumes that depth truncation evaluates \emph{the same model under an alternate runtime configuration}. If an architecture incorporates early-exit classifiers or halting gates that alter behavior upon truncation, this assumption breaks and components become uninterpretable. The architecture examined here uses an \emph{update} gate rather than a halting probability (Equation~\eqref{eq:gate}), satisfying C3.

\paragraph{Condition C4 (Requirement of at Least Two Distinct Blocks).}
The composition component $\Delta_{\text{compose}}$ is defined strictly when $n_b \ge 2$. For single-block recurrent cores, $n_i = N$ and this component is identically zero, reducing the decomposition to two components. This explains why the full decomposition does not transfer to Huginn-0125, even though all other controls transfer seamlessly.

\paragraph{Limit 3 (Calibration Component is a Lower Bound).}
Temperature scaling is a single-parameter family. If miscalibration departs from a pure temperature shift---such as mean logit drift---fitting $T$ captures only the variance explainable by temperature. Therefore, $\Delta_{\text{calib}}$ must be interpreted as a \emph{lower bound} on total miscalibration. Furthermore, it overlaps with the structural components by definition (measured across the same endpoints) and must not be added to their sum.

\paragraph{Limit 4 (Scope of the Negative Control).}
The negative control supports the conclusion that confounders are not artifacts of the measurement protocol. It matches the truncation schedule, temperature fitting routine, and evaluation data, but does \emph{not} match parameter scale, pretraining corpus, or architectural family. It cannot rule out hypotheses involving unobserved variables that co-vary with the depth schedule.

\paragraph{Limit 5 (Intervention Establishes Sufficiency, Not Necessity).}
The controlled intervention in Section~\ref{sec:sampledepth} perturbs exactly one variable against a control branch receiving an identical training budget. This proves the depth schedule is \emph{sufficient} to shift calibration share. It does not prove necessity, nor does it exclude the possibility of alternative interventions producing similar effects.

\paragraph{Condition C5 (Quality Metric Selection).}
The metric $Q$ must exhibit sufficiently low variance to resolve differences between adjacent configurations. On undertrained checkpoints, single-digit generation accuracy exhibits variance that overwhelms the underlying effect size. Consequently, all quantities in this paper are likelihood-based. DCP does not strictly require $Q$ to be a likelihood metric, but mandates that its variance be small relative to effect differences---a condition that must be verified prior to adopting task-level metrics.

\subsection{Component 3: Negative Control on Dense Transformers}
\label{sec:negctrl-method}

The three positive controls characterize how the apparent gap decomposes, but do not establish whether that decomposition is an artifact of the evaluation protocol. Any truncation protocol could potentially induce confounders, for instance if removing layers consistently shifts residual distributions in a uniform direction. DCP therefore requires a negative control: applying identical interventions to layer truncation on standard dense transformers, where the hypothesized confounder mechanisms should not operate. If the measured shares on the negative control differ in sign or magnitude from the target model, the confounders represent genuine properties of the evaluated model rather than measurement artifacts. In dense transformers, repeat controls are unrealizable; hence, the negative control executes prefix, suffix, and calibration sweeps.

\subsection{Component 4: Controlled Training Intervention}
\label{sec:intervention-method}

The preceding components provide correlational evidence: confounders appear in some models and are absent in others. To establish causal attribution, DCP mandates a training intervention modifying \emph{exactly one suspected causal variable} against a control branch receiving an identical step budget without the intervention. In this paper, that variable is the depth schedule: the intervention branch trains with uniformly sampled recurrence depth, while the control branch trains at fixed depth. Both branches are subsequently evaluated using the identical positive controls. This paired design eliminates the competing hypothesis that observed shifts merely reflect additional training steps.

% =====================================================================
\section{Experimental Setup}
\label{sec:setup}

\paragraph{Model Under Study.}
The primary empirical subject of this paper is Sona, a depth-recurrent language model
we trained for mathematical reasoning. Sona is not the method proposed in this paper;
the proposed method is DCP (Section~\ref{sec:method}), and Sona serves as one of five configurations
evaluated under DCP. We utilize our own model as a primary benchmark because DCP requires precise knowledge
of the depth schedule during pretraining---an operational detail seldom released with model weights.

Sona is a 542.8M parameter decoder-only model featuring a two-phase layer stack:
$16$ perception layers, followed by $n_b = 2$ reasoning blocks, each applied $n_i = 4$ times,
yielding a total of $8$ block applications per forward pass. The hidden dimension is $1536$
with $24$ attention heads ($4$ KV heads) and an FFN intermediate dimension of $4352$.
A latent ``thought'' bank of $32$ tokens is carried forward across recurrence steps.
Attention between the thought bank and the textual sequence is restricted:
thought$\rightarrow$thought is unmasked, thought$\rightarrow$text and text$\rightarrow$thought
are blocked, and text$\rightarrow$text is strictly causal; the two streams communicate exclusively
through a causal bridge. Positional encodings utilize rotary position embeddings (RoPE)
\citep{su2024roformer} via a dual-stream variant. Each reasoning block computes an energy-based gate
$g = \sigma\!\left(-(E + \alpha|\Delta E|)/\tau\right)$, applied per-position and per-iteration:
\begin{equation}
  x_{\text{out}} = g \cdot h + (1 - g) \cdot x_{\text{residual}},
  \label{eq:gate}
\end{equation}
establishing that $g$ is an \emph{update} gate governing the proportion of block output written into the residual stream,
rather than a halting probability; the architecture features no early-exit mechanism, departing from adaptive computation
schemes governed by learned halting distributions \citep{graves2016adaptive, banino2021pondernet}.
Equation~\eqref{eq:gate} is transcribed directly from the released implementation codebase, not from design notes.

\paragraph{Checkpoints.}
We evaluate three frozen checkpoints of the same architecture: a pretraining checkpoint at step $34{,}000$,
a downstream checkpoint at step $4{,}000$ of subsequent fine-tuning, and, for replication in Section~\ref{sec:replication},
a second pretraining checkpoint at step $40{,}000$ trained on the decontaminated corpus with repaired document boundaries.
All three checkpoints are undertrained relative to modern compute-optimal small models; absolute task performance is not the object of study.
The downstream checkpoint underwent preference optimization following supervised fine-tuning. Its task accuracy is modest,
which is precisely why all evaluations in this work rely on likelihood metrics rather than discrete generation metrics:
at single-digit accuracy, generative variance completely dominates the structural effects under investigation.

\paragraph{Evaluation Data.}
All measurements are conducted on a held-out set of $200$ problem--solution documents drawn from the English mathematical corpus
used to train Sona, absent from all training splits. This corpus was reconstructed following a decontamination audit detailed separately;
benchmark contamination in mathematical reasoning represents an acknowledged risk \citep{brown2020language, tola23},
so the held-out set was audited against public benchmarks prior to use. Section~\ref{sec:domain} replicates the primary evaluation
on out-of-domain text to verify that conclusions do not depend upon this domain choice.

\paragraph{Metrics.}
The quality metric $Q$ utilized throughout is negative log-likelihood (NLL) under teacher forcing,
evaluated over the solution segment of each document, with next-token accuracy reported as a secondary metric.
Each configuration is evaluated over exactly $125{,}100$ tokens, identical across all experimental sweeps,
so that all cross-condition differences represent paired comparisons over the identical token set.
Likelihood is preferred over generation metrics because, at the performance regime of these checkpoints,
generative sampling variance would obscure the structural effects under analysis.

% =====================================================================
\section{Depth Truncation is Confounded}
\label{sec:depth}

\subsection{Naive Truncation}

Standard depth truncation restricts recurrence to the first $k$ block applications out of the total $8$,
reporting performance as a function of $k$. To ensure learned per-iteration embeddings remain aligned with pretraining,
we preserve the original global iteration index $g_{\text{idx}} = b \cdot n_i + t$ for block $b$ at iteration $t$,
rather than re-indexing from zero.

\begin{figure}[t]
\centering
\includegraphics[width=.72\linewidth]{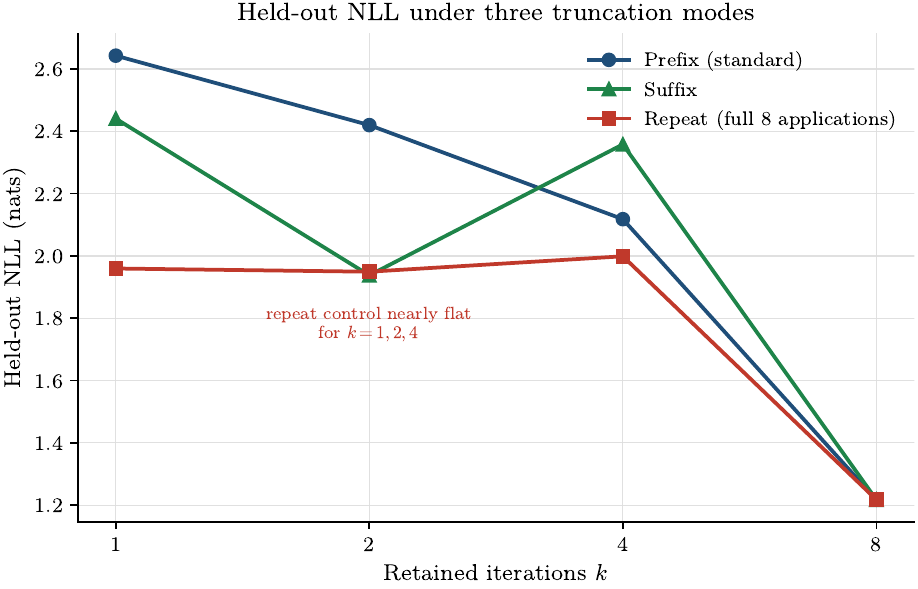}
\caption{Held-out NLL as a function of retained iterations under three truncation modes on the pretraining checkpoint at step $40{,}000$. Prefix truncation drops monotonically. The repeat control unrolls the full $8$ block applications at every $k$, varying only the number of \emph{distinct} iterations; it remains nearly flat across $k = 1, 2, 4$. Complete data in Tables~\ref{tab:prefix} and~\ref{tab:controls}.}
\label{fig:controls}
\end{figure}

Figure~\ref{fig:controls} and Table~\ref{tab:prefix} display this measurement. NLL decreases monotonically
with $k$ across both evaluations, achieving a $1.43$~nat drop on the downstream checkpoint---a figure easily
read as compelling evidence that latent recurrent depth performs substantial reasoning.

\subsection{Applying the Controls}

We apply the three positive controls of DCP (Section~\ref{sec:controls}) to the identical checkpoint and held-out documents
used for naive truncation above. All three intervene strictly at inference time over frozen weights;
hence, all observed differences represent differences in how the same model is probed, not differences in model weights.

\subsection{Decomposition}

Figure~\ref{fig:controls} and Table~\ref{tab:controls} decompose the $1.4251$~nat naive gap ($k{=}1 \rightarrow k{=}8$, prefix, raw).
Because all components are differences across conditions and all shares are ratios of these differences, confidence intervals
computed independently per condition would be misleading: conditions are evaluated on identical documents and are strongly correlated.
We therefore resample document indices \emph{once per bootstrap replication} and propagate that identical sample across all conditions
and derived components ($n = 200$ documents, $5{,}000$ replications). All four component confidence intervals exclude zero:

\begin{itemize}
  \item \textbf{Block applications / residual distribution: $0.6831$~nats,
        $47.9\%$ [$46.6, 49.3$].} Prefix $k{=}1$ ($2.6426$) versus repeat $k{=}1$ ($1.9595$).
        The volume of distinct computation is identical; only the number of block applications differs.
        Nearly half of the apparent depth effect is not depth.
  \item \textbf{Distinct depth within a block: $-0.0392$~nats,
        $-2.8\%$ [$-3.2, -2.4$].} Repeat $k{=}1$ ($1.9595$) versus repeat $k{=}4$ ($1.9987$).
        Four distinct iterations of the first reasoning block are not merely ``no better'' than a single iteration repeated:
        the confidence interval strictly excludes zero on the negative side, establishing that they are reliably
        \emph{worse}, albeit by a modest margin.
  \item \textbf{Inter-block composition: $0.7812$~nats,
        $54.8\%$ [$53.3, 56.3$].} Repeat $k{=}4$ ($1.9987$) versus $k{=}8$ ($1.2173$), the unique configuration where both distinct reasoning blocks
        execute in their trained sequential order.
  \item \textbf{Calibration: $0.3645$~nats, $25.6\%$ [$24.5, 26.6$]}
        (overlapping with the first component). The post-calibration gap is $1.0599$~nats compared to $1.4251$~nats raw,
        and fitted temperature drifts from $1.45$ at $k{=}1$ down to $0.91$ at $k{=}8$ (amplitude $0.57$).
        Here $T$ is fixed to its full-sample optimum, so this interval reflects document sampling rather than uncertainty in $T$ itself.
\end{itemize}

The suffix control reinforces this conclusion. It exhibits non-monotonic behavior: unrolling the four iterations of the second block alone ($2.3579$)
is \emph{worse} than unrolling two of them ($1.9376$), and worse than unrolling four iterations of the first block ($2.1178$).
Iteration count is not the governing causal variable.

\paragraph{Interpretation.} On this model and at this training stage, latent recurrent depth itself contributes negligible measurable quality;
the effect attributed to depth by naive truncation is dominated by residual statistics, readout calibration, and the engagement of a distinct second reasoning block.
We emphasize this is an empirical finding for a specific undertrained checkpoint, not an indictment of recurrent architectures in general;
Section~\ref{sec:huginn} shows that a recurrent model trained with sampled depth behaves entirely differently.
Section~\ref{sec:replication} replicates this full decomposition on a second checkpoint: the ranking of the three primary components is preserved,
while distinct depth remains near zero but flips sign---an observation that leads us to frame our conclusion in a more conservative, robust form.

\subsection{Replication Across Two Independent Checkpoints}
\label{sec:replication}

The decomposition above evaluates a single checkpoint, which is vulnerable to the critique that attribution shares may be idiosyncratic to that model.
We replicated the \emph{entire} protocol (identical held-out set, identical $n = 200$ documents, identical $5{,}000$ bootstrap replications,
and identical CLI flags transcribed directly from original run logs) on a pretraining checkpoint at step $40{,}000$, trained further
on the decontaminated \texttt{en\_math\_v4} corpus with repaired document boundaries.

\begin{figure}[t]
\centering
\includegraphics[width=.82\linewidth]{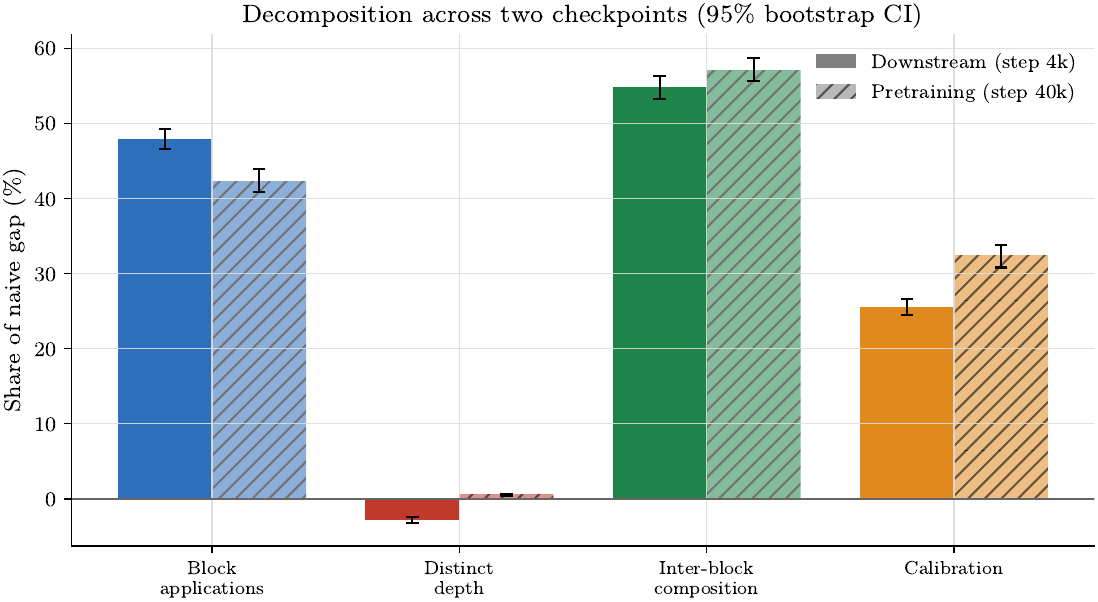}
\caption{Decomposition of the naive gap into four constituent components across two checkpoints using the identical evaluation protocol. Error bars denote $95\%$ bootstrap confidence intervals over $5{,}000$ document resamplings. The distinct depth component is near zero in both checkpoints and reverses sign between them.}
\label{fig:decomposition}
\end{figure}

Figure~\ref{fig:decomposition} and Table~\ref{tab:replication} present both decompositions side by side.
The new checkpoint shares lineage with the pretraining checkpoint in Section~\ref{sec:setup} but diverged in pretraining data later in training,
and is separated from the downstream checkpoint by thousands of training steps. This serves as a test of \emph{robustness} rather than full independent replication.

\paragraph{A Third Checkpoint, Post-Reinforcement Learning.}
The replication above spans two pretraining checkpoints of shared lineage. To evaluate under a wider divergence,
we applied the repeat control to a checkpoint that traversed the complete downstream post-training pipeline:
supervised fine-tuning, preference optimization, and $50$ steps of RLOO reinforcement learning on GSM8K problems.
This checkpoint differs in data distribution, loss formulation, and optimization algorithm; if distinct depth shares were an artifact
of pretraining, they should diverge substantially here.

\begin{table}[t]
\centering
\small
\begin{tabular}{lrrr}
\toprule
Distinct iterations & $1$ & $2$ & $4$ \\
\midrule
NLL, full $8$ applications & $2.4266$ & $2.4270$ & $2.4288$ \\
\bottomrule
\end{tabular}
\caption{Repeat control on the post-reinforcement learning checkpoint. Total block applications are fixed at $8$ across all columns; only the number of \emph{distinct} iterations varies. Increasing from one to four distinct iterations shifts NLL by $+0.0022$~nats (a slight degradation).}
\label{tab:replication-rl}
\end{table}

Table~\ref{tab:replication-rl} presents the results. Four distinct iterations yield no performance gain over a single repeated iteration;
the difference is $+0.0022$~nats towards degradation. Full decomposition on this checkpoint attributes $39.2\%$ to block applications,
$60.9\%$ to inter-block composition, and $-0.14\%$ to distinct depth.
Three checkpoints across three disparate training regimes yield the identical conclusion: under the repeat control, additional distinct iterations
buy no measurable performance in this architecture.

\begin{table}[h]
\centering
\small
\begin{tabular}{lccc}
\toprule
Component & Downstream, Step $4{,}000$ & Pretraining, Step $40{,}000$ & Overlapping CI \\
\midrule
Inter-block composition & $+54.8\%$ [$53.3, 56.3$] & $+57.1\%$ [$55.6, 58.7$] & Yes \\
Block applications        & $+47.9\%$ [$46.6, 49.3$] & $+42.3\%$ [$40.8, 43.9$] & No \\
Calibration            & $+25.6\%$ [$24.5, 26.6$] & $+32.4\%$ [$30.8, 33.8$] & No \\
Distinct depth   & $-2.8\%$  [$-3.2, -2.4$] & $+0.5\%$  [$+0.4, +0.6$] & No \\
\midrule
Naive gap  & $1.4251$~nats & $1.4808$~nats & n/a \\
\bottomrule
\end{tabular}
\caption{Decomposition across two checkpoints under identical protocol. The rank ordering and sign of the three dominant components are strictly preserved; distinct depth remains near zero while reversing sign.}
\label{tab:replication}
\end{table}

\paragraph{Repeat Control.} The repeat control produces matching qualitative results.
Holding total applications constant at $8$ and varying only the number of \emph{distinct} iterations, NLL is
$2.2930$, $2.2931$, and $2.2851$ for $1$, $2$, and $4$ distinct iterations, respectively.
Expanding distinct depth from $1$ to $4$ alters NLL by merely $0.0079$~nats.
On the identical model, increasing \emph{block applications} from $1$ to $8$ while fixing distinct depth to $1$
(prefix $k{=}1$, $2.9202$, versus repeat $k{=}1$, $2.2930$) improves NLL by $0.6272$~nats---nearly an $80\times$ larger shift.

\paragraph{Calibration Step Function Replicates.} Fitted temperatures are $1.511$, $1.477$, and $1.505$ at $k = 1, 2, 4$,
dropping to $1.019$ at $k = 8$; on the downstream checkpoint, they are $1.447$, $1.483$, $1.435$, dropping to $0.906$.
Both profiles remain flat across all truncated depths before dropping sharply to $\approx 1$ strictly at full depth,
matching the diagnostic signature described in Section~\ref{sec:step}.

\paragraph{Non-Replicating Sign and Implications for Confidence Intervals.}
The sign of the distinct depth component reverses: $-2.8\%$ on the downstream checkpoint versus $+0.5\%$ here,
with both confidence intervals strictly excluding zero and not overlapping.

We argue the proper interpretation is not that one measurement is invalid, but that bootstrap intervals \emph{underestimate} true operational uncertainty.
Bootstrap resampling over documents with a frozen model captures only document sampling variance; it fails to capture checkpoint-to-checkpoint variance.
Here, inter-checkpoint variation is $3.3$ percentage points---over twenty times the width of either bootstrap interval.
In absolute magnitude, both values are negligible: $-0.0392$ and $+0.0079$~nats out of an overall gap of $\approx 1.45$~nats.

We therefore frame our conclusion in a more robust form: across checkpoints, the distinct depth component falls within $[-3\%, +1\%]$,
indistinguishable from zero, with an unstable sign. This claim is more modest than ``distinct depth harms performance'',
but does not depend upon fragile details and is the sole claim supported by the data.

\paragraph{A Note on Naive Interpretation.} On the newer checkpoint, automated evaluation routines would conclude that ``recurrence works''
across all three regimes, because NLL drops monotonically with $k$ under prefix truncation ($2.9202 \rightarrow 2.6338 \rightarrow 2.3937 \rightarrow 1.4390$).
DCP decomposition reveals that the share attributable to distinct depth is a meager $0.5\%$.
The gulf between these two interpretations constitutes the core thesis of this paper.

\subsection{The Gate Learns a Non-Degenerate Allocation Schedule}

Although distinct depth fails to improve quality, the update gate in Equation~\eqref{eq:gate} is non-degenerate.
Table~\ref{tab:gates} (Appendix) reports gate distributions per iteration across $60$ held-out documents.
The gate regularizer targets a global mean of $0.5$; the model satisfies this globally while learning a monotonically increasing schedule,
updating more aggressively in later iterations, with substantial within-iteration variance.

\subsection{What a Depth Value Head Can ``See''}
\label{sec:probe}

A natural application for depth-recurrent architectures is depth-wise credit assignment: attaching a value head at each iteration
and utilizing its incremental gain as an advantage estimate. We test whether the informational signal required by such a head exists.
For each held-out document, we construct a corrupted counterpart (preserving the reasoning trace but perturbing the final answer),
extract latent representations at each iteration, and train a cross-validated linear probe to discriminate valid from corrupted solutions.

Probe AUC remains near chance and \emph{degrades} with depth:
$0.549 \rightarrow 0.525$ (pretraining) and $0.548 \rightarrow 0.535$ (fine-tuning) from iteration $0$ to $7$.
This is an architectural consequence rather than a training failure: the input to the verifier is
$\text{thought\_state} + W\,\overline{x}_{\text{prefix}}$, where $\overline{x}_{\text{prefix}}$ denotes the \emph{mean} across causal prefix representations,
and the thought bank cannot attend into the textual sequence. A perturbation in a single answer token shifts a multi-hundred-token average negligibly.
Consequently, in this architecture, any depth value head must hook into textual stream representations rather than the latent thought bank.

% =====================================================================
\section{Negative Control: Layer Truncation on Standard Transformers}
\label{sec:negctrl}

This section executes the third component of DCP (Section~\ref{sec:negctrl-method}). The potential failure mode
probed here is not a vulnerability of the model, but a potential artifact of \emph{the measurement protocol itself}:
any truncation procedure could conceivably induce confounders, for instance if removing compute systematically
shifts residual stream distributions in a uniform direction. Were this the case, the two confounders quantified
in Section~\ref{sec:depth} would reveal nothing about Sona, reflecting merely an artifact of the evaluation suite.

The negative control discriminates between these two hypotheses, which constitutes its primary advantage over reporting
positive controls alone: it provides an empirical baseline where the hypothesized confounder mechanism should not operate;
hence, any non-zero share observed in this group must be attributed to measurement artifacts. Specifically, if the confounders
in Section~\ref{sec:controls} were measurement artifacts rather than intrinsic properties of recurrent depth, they should equally
manifest when applying identical interventions to layer truncation in standard transformers---the exact evaluation widely
employed in layer-pruning and early-exit literature. We therefore replicate the protocol across two public models, truncating
to the first $k$ layers out of $L$ (\emph{prefix}) and, as a control, executing the first $k$ layers and repeating the $k$-th
layer until completing $L$ applications (\emph{repeat}).

\paragraph{The Two Components Do Not Transfer Identically.}
The preliminary negative control ran without temperature recalibration, informing solely the application count component.
Re-evaluating with temperature scaling reveals a fundamental divergence between the two confounders.

The \emph{block application} component reverses sign in dense networks, as theoretically predicted:
$-31.7\%$ on Qwen2.5-Math-1.5B and $-21.2\%$ on Qwen3-1.7B. This effect is unique to weight-sharing architectures
and does not transfer to dense models.

The \emph{calibration} component, by contrast, transfers robustly. On Qwen2.5-Math-1.5B evaluated on general text where the model
is well-calibrated ($T = 1.03$ at full depth), fitted temperature drifts from $1.91$ at $k = 4/28$ to $1.03$ at full depth,
accounting for $16.5\%$ of the apparent gap. The baseline $T = 1.03$ at full depth is the crucial sanity check: it confirms
the model is well-calibrated on that domain \emph{prior} to truncation, ensuring all subsequent drift is induced by layer removal
rather than domain shift.

We verified this condition by evaluating on specialized mathematics text, where the identical model exhibits $T = 1.12$
at full depth and a calibration share of only $8.0\%$. The divergence between domains demonstrates why evaluation data must be chosen
where the baseline model is well-calibrated: evaluating on a misaligned domain conflates background drift, paradoxically
\emph{attenuating} the measured calibration share. On Qwen3-1.7B, full-depth temperature is $1.29$ even on general text;
the baseline calibration condition is imperfectly satisfied, so we treat its $28.9\%$ calibration share as secondary evidence.

The implications for the broader literature are substantial. The calibration confounder is not idiosyncratic to recurrent models
trained at fixed depth; it manifests fully when pruning layers in standard dense transformers. Studies interpreting quality curves
over retained layer counts without recalibration \citep{gromov2025unreasonable, men2025shortgpt, wang2025fewer, shi2026understanding}
report a quantity contaminated by this component. We do not claim their empirical findings are invalid; we show that their reported metric
conflates two distinct mechanisms that our control suite successfully disentangles.

\paragraph{Sanity Checks.} Because custom unrolling bypasses standard library forward passes, we verified that at $k = L$,
manual layer unrolling bitwise reproduces the standard reference forward pass. Across both models, the maximum absolute logit divergence
is $0$ with $100\%$ argmax consensus, confirming the exactness of our attention masking and rotary embedding implementations.

\begin{figure}[t]
\centering
\includegraphics[width=.8\linewidth]{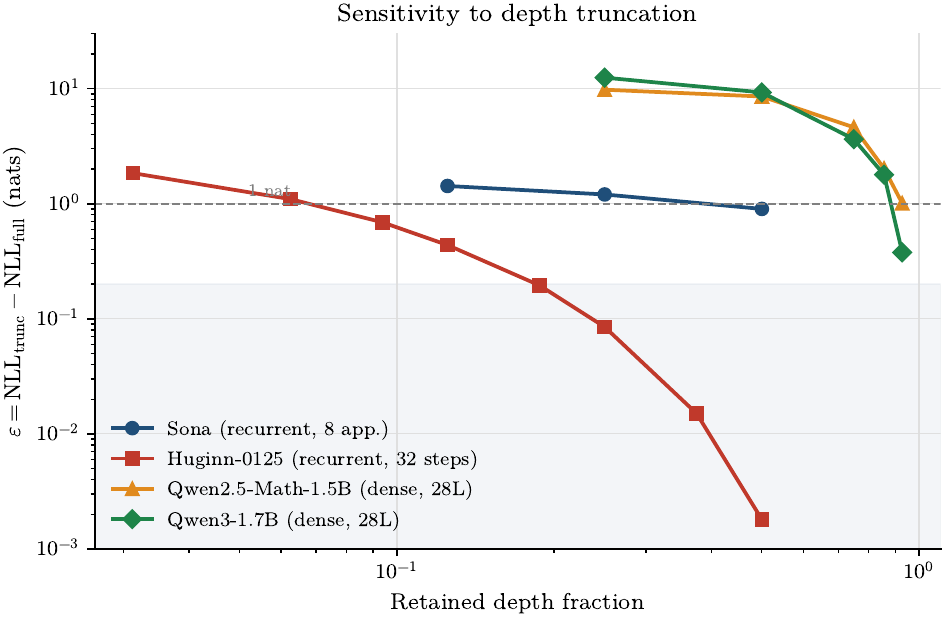}
\caption{$\varepsilon = \text{NLL}(\text{truncated}) - \text{NLL}(\text{full})$, computed within each respective model to enable cross-tokenizer comparison. Log--log scale. The full-depth point is omitted since $\varepsilon = 0$ by definition. The two recurrent models remain below $0.2$~nats across most of the depth range; the two dense transformers exceed $1$~nat even at the mildest measurable truncation. Data in Tables~\ref{tab:negctrl} and~\ref{tab:huginn}.}
\label{fig:elasticity}
\end{figure}

\begin{table}[t]
\centering
\begin{tabular}{lccc}
\toprule
& Sona (Recurrent) & Qwen2.5-Math-1.5B & Qwen3-1.7B \\
\midrule
Naive gap (nats) & $1.4253$ & $9.7853$ & $12.4609$ \\
Application count share & $+47.9\%$ & $-35.9\%$ & $-39.9\%$ \\
Calibration share & $+25.6\%$ & $+3.3\%$ & $+14.5\%$ \\
\bottomrule
\end{tabular}
\caption{Cross-architecture decomposition. The ``application count'' confounder reverses sign on standard transformers, and the calibration confounder is substantially attenuated. Both represent structural characteristics of weight-tied recurrent depth.}
\label{tab:negctrl-decomp}
\end{table}

\paragraph{Results.} Neither confounder transfers unchanged. The application count share turns negative across both dense models
($-35.9\%$ and $-39.9\%$): repeating a layer degrades standard transformers across all depths, except when within two layers
of the full stack. The calibration share is $3.3\%$ and $14.5\%$, compared to $25.6\%$ on the depth-recurrent architecture.

\paragraph{Mechanism.} In a standard transformer, each layer is an independent function optimized strictly for its specific index
within the network stack; repeating an intermediate layer shifts the residual stream severely out of distribution: at $k = 14/28$,
repetition drives perplexity from $1.3 \times 10^4$ to $3.7 \times 10^6$ on Qwen2.5-Math. In a depth-recurrent model, by contrast,
the block is explicitly designed for iterative execution, rendering application count a free variable that can be held fixed while manipulating
distinct compute. Consequently, the controls in Section~\ref{sec:controls} are not generic prescriptions for all depth truncation;
they are tailored diagnostic controls for recurrent models, and this experiment delineates their valid domain of application.

\paragraph{Caveat.} Absolute NLL values are not cross-comparable across architectures: truncating a standard transformer to half-depth
induces catastrophic failure (perplexity $>10^4$), whereas the recurrent model degrades smoothly. The comparable quantities are the sign
and proportional attribution shares of each decomposed component, not the absolute magnitude of the performance drop.

% =====================================================================
\section{Vulnerability Analysis: Training at Fixed Depth}
\label{sec:huginn}

Huginn-0125 \citep{geiping2026scaling} is a 3.5B parameter depth-recurrent model comprising $2$ prefix layers,
a $4$-layer recurrent core, and $2$ suffix layers. Crucially, its recurrence depth was \emph{sampled during pretraining}
(\texttt{mean\_recurrence}$=32$, following a Poisson--log-normal distribution), whereas Sona was trained exclusively at fixed depth $8$.
Consequently, Huginn's readout head observed residual states originating from diverse recurrence depths rather than a single fixed depth,
providing a direct empirical test of the generative mechanism behind calibration drift.

\paragraph{What Does Not Transfer, and Why.} Huginn's recurrent core consists of a \emph{single weight-tied block};
hence, unrolling it for $k$ iterations \emph{is} repeating it: application count and distinct compute are identical variables.
The structural decomposition in Section~\ref{sec:controls} is therefore \emph{formally undefined} here, rather than merely unmeasured.
Only the calibration control transfers, which we report below.

\paragraph{Results.} Fitted temperature remains tightly bounded in $[0.94, 0.99]$ across a $32\times$ depth range (drift amplitude $0.045$),
and recalibration accounts for merely $-0.6\%$ of the $1.84$~nat depth gap. By contrast, the fixed-depth model exhibits a drift amplitude of $0.57$
across an $8\times$ range and a $25.6\%$ calibration share---an order of magnitude larger drift.

This finding isolates the underlying mechanism, which Section~\ref{sec:sampledepth} verifies via a controlled training intervention.
The confounder is linked specifically to \emph{training at fixed depth}, not recurrent depth per se: a readout head trained on residuals
from a single recurrence depth suffers calibration drift when evaluated at other depths. This directly suggests a practical remedy:
randomly sampling recurrence depth during training. It likewise explains why standard transformers (also trained at fixed layer counts)
in Section~\ref{sec:negctrl} retain modest calibration effects ($3.3\%$ and $14.5\%$), whereas the depth-sampled model shows none.

\paragraph{Two Ancillary Observations.} Huginn's performance saturates around $r \approx 16$
($1.2899 \rightarrow 1.2888$ from $r=16$ to $r=32$, $\Delta = 0.0009$~nats), despite being trained with a mean recurrence of $32$.
Moreover, it demonstrates a substantial, monotonic, unconfounded depth effect of $1.83$~nats.
The appropriate takeaway from Section~\ref{sec:controls} is therefore not that ``recurrent depth cannot help''
(a well-trained recurrent model unambiguously leverages its depth), but that \emph{this specific model},
trained at fixed depth and undertrained, fails to utilize its own depth.

\paragraph{Caveat.} Huginn and Sona differ in parameter scale (3.5B vs.\ 542.8M), tokenizers, and pretraining corpora;
hence, this provides mechanistic evidence rather than a controlled experiment. The fully controlled counterpart---training the identical
architecture twice under fixed versus sampled depth---is executed in Section~\ref{sec:sampledepth}.

\paragraph{Pattern Across Four Models.} Table~\ref{tab:fourmodel} synthesizes all calibration evaluations across this paper.
Across four models, calibration share tracks whether training employed fixed or sampled depth: two standard transformers and one recurrent model,
all trained at a single fixed depth, yield calibration shares between $3.3\%$ and $25.6\%$; the single model trained with sampled depth yields $-0.6\%$.

\begin{table}[h]
\centering
\begin{tabular}{llcc}
\toprule
Model & Training Depth Schedule & Temperature Range & Calibration Share \\
\midrule
Sona (542.8M, recurrent)         & Fixed, $8$        & $0.91$--$1.48$ & $25.6\%$ \\
Qwen3-1.7B                  & Fixed, $28$ layers   & $1.45$--$2.83$ & $14.5\%$ \\
Qwen2.5-Math-1.5B           & Fixed, $28$ layers   & $1.12$--$1.42$ & $3.3\%$ \\
Huginn-0125 (3.6B, recurrent) & \textbf{Sampled}, mean $32$ & $0.94$--$0.99$ & $\mathbf{-0.6\%}$ \\
\bottomrule
\end{tabular}
\caption{Calibration confounder tracks depth variation during training, invariant to architectural family and parameter scale.}
\label{tab:fourmodel}
\end{table}

% =====================================================================
\section{Dynamics of Recurrence}
\label{sec:dynamics}

Section~\ref{sec:controls} demonstrated that additional distinct iterations fail to improve quality, but did not reveal why.
This section directly tracks the \emph{latent state trajectories} through successive block applications.
All three dynamical measurements require only forward passes without retraining, enabling evaluation across all models in this study.

\paragraph{Gate Magnitudes Do Not Measure What They Seem.} Table~\ref{tab:gates} reports update gate values $\gamma^{(g)}$,
increasing monotonically from $0.34$ to $0.70$. However, $\gamma$ governs only the \emph{mixture ratio} between block output and residual stream;
if block output is aligned with incoming residual states, the hidden state remains nearly invariant even with a wide-open gate.
The proper metric is relative update magnitude $\|x^{(g)} - x^{(g-1)}\| / \|x^{(g-1)}\|$, which on the fixed-depth recurrent model
is \emph{non-monotonic}: $0.26$, $0.35$, $0.40$, $0.33$, $0.29$, $0.39$, $0.52$, $0.41$. The gate schedule and true update trajectory tell divergent stories.

\paragraph{Iterations Compute Distinct Transformations.} Stacking normalized updates $\Delta x^{(g)}$ into a matrix and evaluating
the participation ratio $(\sum_i \sigma_i)^2 / \sum_i \sigma_i^2$ of singular values provides a soft estimate of the subspace dimensionality spanned.
The recurrent model achieves $6.24$ out of a maximum $8$ ($78\%$), and $\cos(\Delta x^{(g)}, \Delta x^{(1)})$ drops from $0.55$ to $0.06$ by the fourth iteration.
The iterations are non-redundant: they execute geometrically distinct transformations.

Juxtaposed with the repeat control, this sharpens our primary finding rather than explaining it away. The model computes eight largely non-redundant
transformations, yet replacing four distinct iterations with a single repeated iteration alters NLL by merely $0.0079$~nats.
Computational diversity \emph{exists internally} but is \emph{not decoded by the readout}.

\paragraph{Perturbation Response.} We inject Gaussian noise with norm matching $10\%$ of state norm into an intermediate application,
tracking $\|\delta^{(g)}\| / \|\delta^{(0)}\|$ across subsequent applications. This represents the sole metric among the three directly
comparable between recurrent and dense models, as it is scale-invariant: in dense networks, early layers operate on unnormalized embeddings
where $\|\Delta x\|/\|x\|$ reaches $25$--$33$, on a completely different scale from recurrent steps.

\begin{table}[h]
\centering
\small
\begin{tabular}{llcc}
\toprule
Model & Type & Per-Iteration Noise Ratio & Participation Ratio \\
\midrule
Qwen2.5-Math-1.5B            & Dense, $28$ layers   & $1.011$ & $90\%$ \\
Sona $+1{,}000$ sampled steps    & Recurrent, $8$ applications     & $1.053$ & $69\%$ \\
Sona pretraining $40$k   & Recurrent, $8$ applications     & $1.067$ & $78\%$ \\
Sona $+1{,}000$ fixed steps    & Recurrent, $8$ applications     & $1.069$ & $76\%$ \\
Qwen3-1.7B                   & Dense, $28$ layers   & $1.060$ & $90\%$ \\
\midrule
\textbf{Huginn-0125}         & Recurrent, $32$ steps   & $\mathbf{0.892}$ & $95\%$ \\
\bottomrule
\end{tabular}
\caption{Perturbation response. A ratio below $1$ indicates perturbation attenuation and contraction towards a fixed point; above $1$ indicates perturbation amplification. The first four configurations expand perturbations at comparable rates, irrespective of recurrent or dense architecture. Huginn is the sole exception. Evaluated on $100$ documents for Sona, $40$--$60$ for Huginn and dense transformers.}
\label{tab:dynamics}
\end{table}

\begin{figure}[t]
\centering
\includegraphics[width=\linewidth]{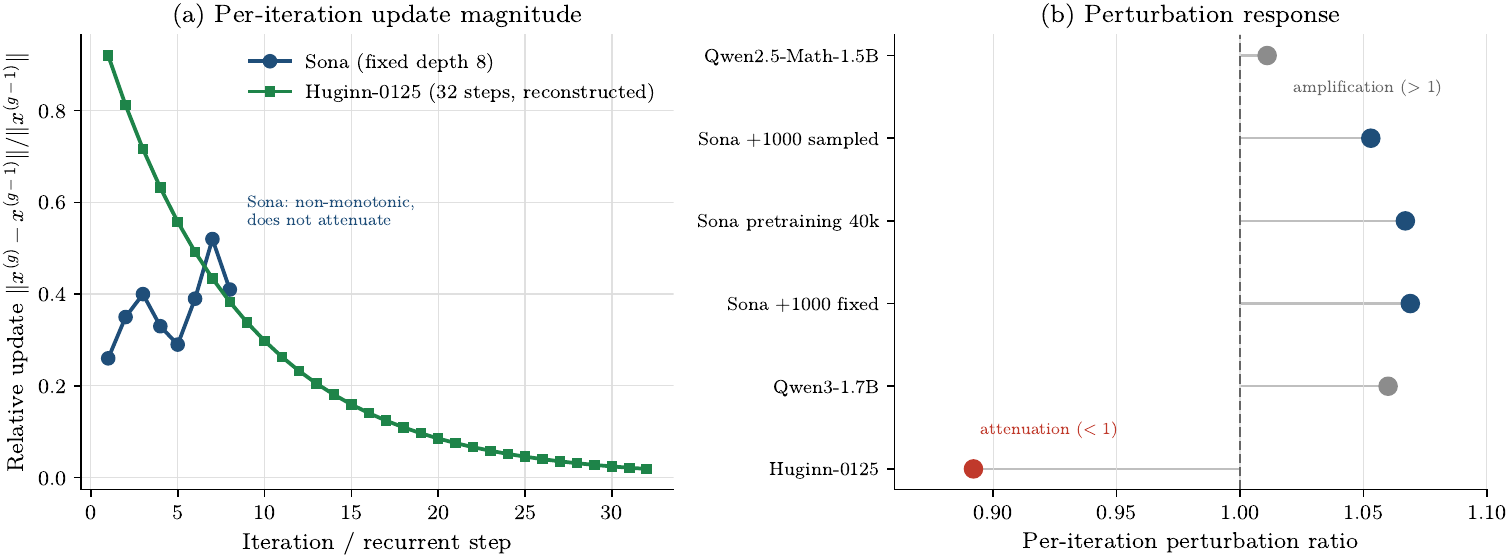}
\caption{\textbf{(a)} Per-iteration update norms, weight-tied models only; dense models differ in scale. In Huginn, update norms decay exponentially from $0.92$ to $0.019$; in Sona, they do not attenuate. \textbf{(b)} Perturbation response on log scale across all five configurations. Only Huginn lies below the unity threshold.}
\label{fig:dynamics}
\end{figure}

\paragraph{Contractivity Does Not Track Depth Schedule.} Prior to evaluating dense transformers, a natural hypothesis was to attribute
contractivity to sampled-depth training, since Huginn possesses both traits. Two empirical measurements refute this reading.

First, the depth-sampling intervention in Section~\ref{sec:sampledepth}, which reduces calibration share from $29.9\%$ to $5.1\%$ on this identical model,
shifts the noise ratio only from $1.069$ to $1.053$ (remaining $>1$). Depth sampling remedies calibration drift but \emph{fails} to alter operator dynamics;
the two mechanisms are orthogonal.

Second, both dense transformers also amplify perturbations ($1.011$ and $1.060$), with Qwen3-1.7B compounding perturbations over $80\times$ across $28$ layers.
Perturbation amplification is a pervasive characteristic of deep layer stacks, not an idiosyncratic defect of our evaluated model:
it falls squarely within the range of fully converged production transformers.

The accurate characterization is therefore that \emph{Huginn learned a contraction mapping while the remaining four configurations did not},
and we \emph{cannot} attribute this dynamical property to the depth schedule. Potential explanatory factors include model scale (3.6B vs.\ 543M and 1.5--1.7B),
pretraining token volume, or recurrent core architecture. We report the empirical observation while leaving the causal origin open.

\paragraph{Explaining Huginn's Empirical Resilience.} Section~\ref{sec:negctrl} observed that truncating Huginn to half-depth incurs merely $0.0018$~nats,
without an obvious explanation. The dynamical data provides a direct answer: beyond iteration $16$, update magnitudes fall below $0.03$.
Truncating that tail is virtually free because negligible computation is discarded. In the fixed-depth model, by contrast, truncation \emph{does}
discard substantial computation, yet performance still fails to improve.

\paragraph{Limitations.} These five configurations differ simultaneously across parameter scale, tokenizers, pretraining corpora, and optimization recipes;
hence, cross-model comparisons remain observational. The sole strictly interventional comparison is the pair of $+1{,}000$ steps (fixed vs.\ sampled depth),
initiated from the identical checkpoint with matching data and random seeds, which demonstrates that depth schedule is \emph{not} the governing variable for contractivity.
We also do not know how this coefficient evolves with extended training: our evaluated checkpoint reached $40{,}000$ out of $110{,}000$ planned steps.
Because this measurement takes roughly two minutes, it represents an inexpensive diagnostic metric to track across future training milestones.

% =====================================================================
% =====================================================================
\section{Direct Verification and Generality}
\label{sec:intervention}

The preceding sections established a correlational pattern: the calibration confounder manifests in fixed-depth models
and vanishes in sampled-depth models. This section executes the fourth component of DCP (Section~\ref{sec:intervention-method})---a controlled
training intervention---to establish causal attribution, accompanied by two tests of empirical generality.

\subsection{Calibration Drift is a Step Function, Not a Gradual Drift}
\label{sec:step}

\begin{figure}[t]
\centering
\includegraphics[width=.72\linewidth]{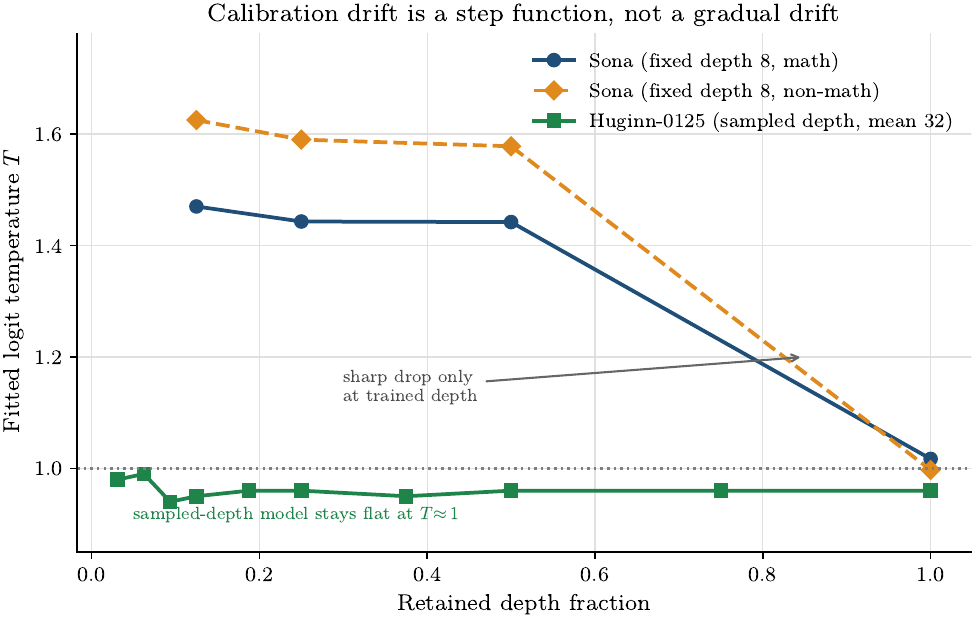}
\caption{Fitted logit temperature as a function of retained depth fraction. In fixed-depth models, $T$ remains nearly constant across all truncated depths before dropping sharply to $\approx 1$ strictly at full depth. In sampled-depth models, $T$ hovers around $1$ across the entire range. Data in Table~\ref{tab:tstep}.}
\label{fig:tstep}
\end{figure}

Figure~\ref{fig:tstep} demonstrates that fitted temperature does not drift gradually with depth. It remains virtually constant
in the range $1.38$--$1.63$ across all truncated depths, before dropping sharply to $\approx 1.00$ strictly at full depth.
Because $T = 1.0$ indicates that \emph{no calibration adjustment is needed}, the proper reading is: the readout head is calibrated
accurately at the precise depth it was trained upon, and miscalibrated to an approximately uniform degree at all other depths.
This pattern holds across all four evaluated configurations, including out-of-domain text.

\subsection{Confounders Are Not Confined to Mathematical Text}
\label{sec:domain}

Replicating the entire evaluation across $200$ non-mathematical documents (a FineWeb slice of the identical corpus)
yields a calibration share of $41.4\%$, compared to $29.9\%$ on mathematical text on the same checkpoint.
The confounder not only persists out-of-domain, but is \emph{amplified}; hence, it is not an idiosyncratic artifact of mathematical syntax.

\subsection{Depth Sampling Intervention}
\label{sec:sampledepth}

To causally verify the generative mechanism identified in Section~\ref{sec:huginn} on the identical model exhibiting confounding,
we trained three continued branches from the same checkpoint, using matching data, seeds, step budget ($1{,}000$ steps),
and constant learning rate. The sole divergence between branches is the pretraining depth schedule.

\begin{figure}[t]
\centering
\includegraphics[width=\linewidth]{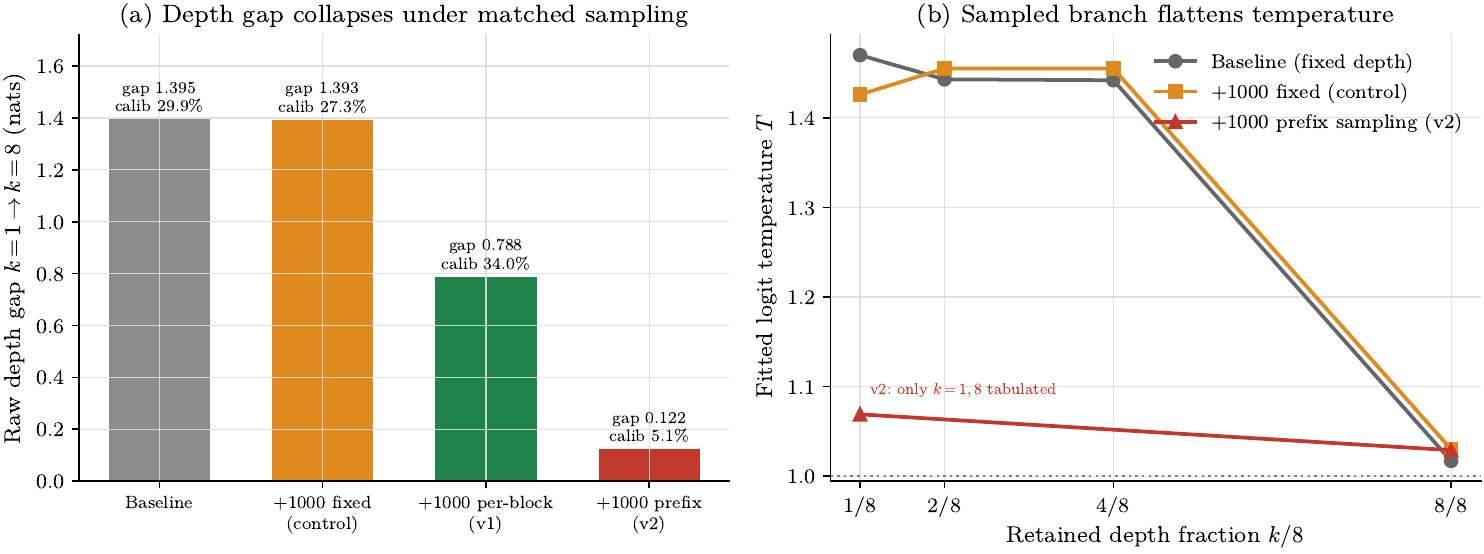}
\caption{Depth sampling intervention. Left: NLL versus retained iterations. Right: Fitted temperature. The sampled branch flattens the temperature profile; the control branch receiving the identical training budget at fixed depth does not.}
\label{fig:intervention}
\end{figure}

\paragraph{Sampling Distribution Must Match Evaluated Deployment Space.}
Our first attempt sampled recurrence iterations \emph{within each block independently}; hence, the model always executed
both blocks, and total block applications strictly took values in $\{2,4,6,8\}$.
During evaluation, prefix truncation with $k \le 4$ executes exclusively the first block: configurations evaluated at $k = 1,2,4$
\emph{never occurred} during pretraining. As theoretically predicted, calibration share did not attenuate (Figure~\ref{fig:intervention}
and Table~\ref{tab:intervention}, row v1). Our second attempt sampled $k \sim \mathcal{U}\{1..8\}$ \emph{total block applications},
executing the exact prefix unrolling of length $k$---matching the configuration family probed during evaluation.

\begin{table}[t]
\centering
\small
\begin{tabular}{lccccc}
\toprule
& Raw Gap & Calibration Share & $T$ at $k{=}1$ & $T$ at $k{=}8$ & $T$ Drift Amplitude \\
\midrule
Baseline                                & $1.3955$ & $29.9\%$ & $1.470$ & $1.017$ & $0.453$ \\
$+1{,}000$ steps, fixed (control)       & $1.3927$ & $27.3\%$ & $1.426$ & $1.030$ & $0.429$ \\
$+1{,}000$ steps, per-block sampling (v1) & $0.7877$ & $34.0\%$ & $1.406$ & $1.028$ & $0.378$ \\
$+1{,}000$ steps, prefix sampling (v2)  & $\mathbf{0.1224}$ & $\mathbf{5.1\%}$ & $\mathbf{1.069}$ & $1.029$ & $\mathbf{0.040}$ \\
\bottomrule
\end{tabular}
\caption{Depth sampling intervention, $n = 200$ documents. Branch v2 samples across the deployment schedule probed during evaluation and collapses the calibration confounder; the control branch, receiving an identical $1{,}000$ steps, remains virtually unchanged from baseline.}
\label{tab:intervention}
\end{table}

\paragraph{Results.} Branch v2 precisely reproduces the diagnostic signature of sampled-depth models: fitted temperature
remains bounded in $[1.029, 1.069]$ across the entire range (drift amplitude $0.040$, matching the $0.045$ of Huginn-0125
and compared to $0.453$ in this model prior to intervention). Calibration share collapses from $29.9\%$ to $5.1\%$.
Because the control branch received the identical $1{,}000$ training steps yet shifted only from $29.9\%$ to $27.3\%$,
the causal driver is unambiguously the depth schedule rather than additional optimization.

This represents causal evidence on the identical model, identical data, and identical compute budget, perturbing a single operational variable.
It elevates the finding in Section~\ref{sec:huginn} from a cross-model observation to a verified causal relationship.

\paragraph{Two Trade-Offs, and One Implication.} First, full-depth performance degrades: NLL at $k=8$ increases from $1.4421$ to $1.5797$.
Multi-depth training trades peak performance for robust generalizability across depths, as theoretically expected. Second, the apparent depth gap
largely vanishes ($1.3955 \rightarrow 0.1224$~nats): post-intervention, executing one application trails eight applications by merely $0.12$~nats.
Coupled with the finding in Section~\ref{sec:controls} that distinct depth contributes $-2.8\%$, this reinforces the conclusion
that recurrence in this architecture buys negligible performance: calibration confounders can be remedied, but the residual effect size is minimal.

\paragraph{Limitations.} The intervention spanned $1{,}000$ steps on a single model, single seed, initiated from an undertrained checkpoint.
We have not established whether training with sampled depth from initialization avoids the peak performance penalty.

% =====================================================================
\section{Benchmarking the Protocol on Known Standards}
\label{sec:calibration-grid}

All preceding results applied DCP to models where the ground-truth attribution was \emph{unknown a priori}.
This leaves open a critical methodological challenge: what guarantees the protocol measures ground truth rather than generating internally consistent numbers?
This section addresses this by constructing three models where the ground truth is known \emph{by construction},
verifying whether DCP faithfully recovers the known true states.

\paragraph{Experimental Design.}
Three branches share identical architecture, corpora, random seeds, and training step budgets.
The sole manipulated variable is the pretraining depth schedule:

\begin{center}
\small
\begin{tabular}{llll}
\toprule
Branch & Pretraining Depth Schedule & Readout Observed $k \in$ & Ground Truth \\
\midrule
A & Fixed at $8$              & $\{8\}$          & Confounded (YES) \\
B & Sampled $\mathcal{U}\{1..8\}$ & $\{1,\dots,8\}$ & Unconfounded (NO) \\
C & Sampled $\mathcal{U}\{5..8\}$ & $\{5,\dots,8\}$ & Confounded at $k = 1,2,4$ (YES) \\
\bottomrule
\end{tabular}
\end{center}

Branches A and B test \emph{sensitivity}: does the protocol detect the confounder when present by construction,
and remain silent when absent? Branch C tests \emph{specificity}, representing the most discriminative test.
It disentangles two competing hypotheses that Branch B alone cannot separate: ``whether depth was sampled'' versus
``whether the readout has observed the specific runtime configurations being evaluated''. If our hypothesized mechanism
holds, Branch C must exhibit confounding at depths outside its training support, despite being trained under variable depth.

\paragraph{Pre-Registered Predictions.}
The mechanistic theory makes a sharp structural prediction: in Branch C, fitted temperature must hover near $1.0$ at $k = 5,\dots,8$
and drift at $k = 1,2,4$---namely, \emph{the step discontinuity must align with the sampling support boundary}.
This is a prediction regarding the \emph{spatial location} of a discontinuity, rather than a loose difference in signs.
That boundary is not fitted to empirical data; it is determined by the pretraining schedule prior to evaluation.
The decision threshold for calibration share was fixed at $5\%$, midway between $-0.6\%$ on Huginn-0125 and $25.6\%$
on fixed-depth models. All three predictions and thresholds were hard-coded into synthesis scripts prior to executing the branches.

\begin{figure}[t]
\centering
\includegraphics[width=\linewidth]{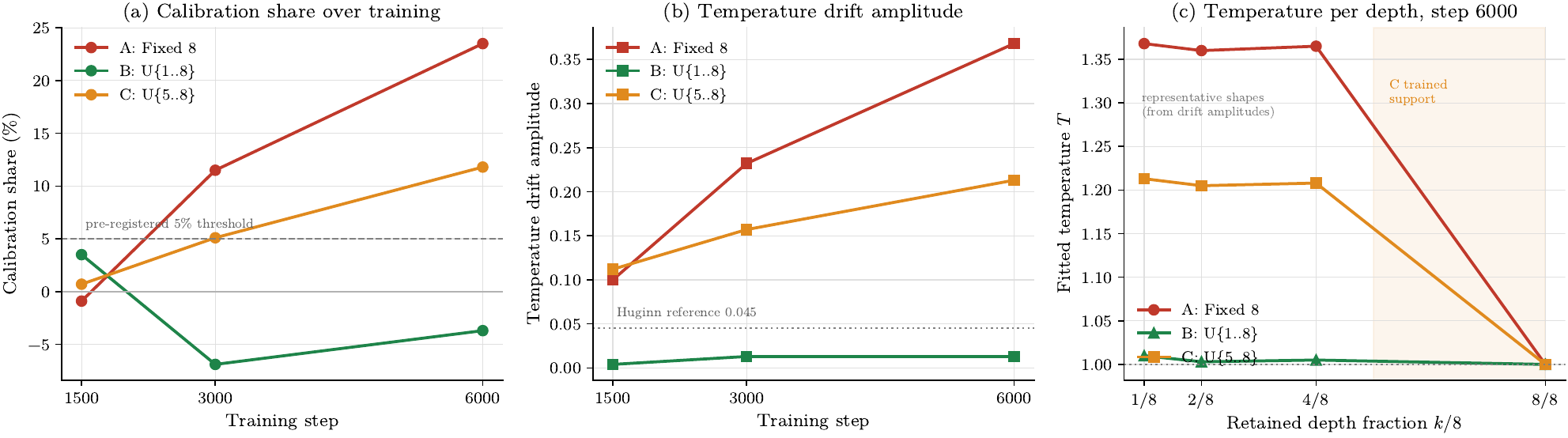}
\caption{Calibration grid across three models differing in a single operational variable: pretraining depth schedule. \textbf{(a)} Calibration share over training budget. Branch A steadily ascends, Branch B hovers around zero, Branch C lies in between. Confounders accumulate with training rather than being warm-up artifacts. \textbf{(b)} Temperature drift amplitudes mirror this ranking; Branch B matches the $0.045$ observed in Huginn-0125. \textbf{(c)} Fitted temperature per depth at step $6{,}000$. Branch A exhibits a step function, dropping to $\approx 1$ strictly at training depth. Branch B remains flat at $\approx 1$. Branch C drops as it enters its trained support (shaded region). Because the grid samples $k \in \{1,2,4,8\}$, the exact boundary transition profile is unresolved.}
\label{fig:grid}
\end{figure}

\paragraph{Results.}
Table~\ref{tab:grid} and Figure~\ref{fig:grid} present the complete calibration grid. All three branches strictly conform to pre-registered predictions.

\begin{table}[t]
\centering
\small
\begin{tabular}{llrrrrrr}
\toprule
Branch & Depth Schedule & Step & $\Delta_{\text{raw}}$ & Calibration & $T$ Drift
      & Application & Distinct \\
\midrule
A & Fixed $8$ & $1500$ & $0.732$ & $-0.9\%$ & $0.100$ & $40.3\%$ & $+0.15\%$ \\
A & Fixed $8$ & $3000$ & $0.926$ & $11.5\%$ & $0.232$ & $38.6\%$ & $-1.69\%$ \\
A & Fixed $8$ & $6000$ & $1.298$ & $\mathbf{23.5\%}$ & $0.368$ & $42.7\%$ & $-4.81\%$ \\
\midrule
B & $\mathcal{U}\{1..8\}$ & $1500$ & $0.083$ & $3.5\%$  & $0.004$ & $22.0\%$ & $+3.65\%$ \\
B & $\mathcal{U}\{1..8\}$ & $3000$ & $0.069$ & $-6.9\%$ & $0.013$ & $53.3\%$ & $+6.59\%$ \\
B & $\mathcal{U}\{1..8\}$ & $6000$ & $0.057$ & $\mathbf{-3.7\%}$ & $0.013$ & $41.8\%$ & $+4.80\%$ \\
\midrule
C & $\mathcal{U}\{5..8\}$ & $1500$ & $0.673$ & $0.7\%$  & $0.112$ & $55.3\%$ & $-1.05\%$ \\
C & $\mathcal{U}\{5..8\}$ & $3000$ & $0.965$ & $5.1\%$  & $0.157$ & $70.2\%$ & $-2.15\%$ \\
C & $\mathcal{U}\{5..8\}$ & $6000$ & $1.230$ & $\mathbf{11.8\%}$ & $0.213$ & $72.8\%$ & $-1.60\%$ \\
\bottomrule
\end{tabular}
\caption{Calibration grid. Three branches share identical architecture, corpora, seeds, and training step budgets, differing strictly in depth schedule. The pre-registered $5\%$ threshold for calibration share was established prior to run execution. At maximum training budget: Branch A yields $23.5\%$ (predicted YES), Branch B yields $-3.7\%$ (predicted NO), Branch C yields $11.8\%$ (predicted YES).}
\label{tab:grid}
\end{table}

Table~\ref{tab:gridci} decomposes the four components with paired $95\%$ bootstrap confidence intervals at step $6{,}000$. We pair the resampling: re-drawing documents once per bootstrap replicate ($n = 200$ documents, $5{,}000$ iterations) and propagating that exact sample across all conditions. All twelve confidence intervals exclude zero.

\begin{table}[t]
\centering
\small
\begin{tabular}{llrr}
\toprule
Branch & Component & Share & $95\%$ CI \\
\midrule
\multirow{4}{*}{A: Fixed $8$}
  & Block applications      & $42.74\%$ & $[41.45, 44.06]$ \\
  & Distinct iterations     & $\mathbf{-4.81\%}$ & $[-5.21, -4.40]$ \\
  & Inter-block composition & $62.07\%$ & $[60.53, 63.56]$ \\
  & Calibration drift       & $22.45\%$ & $[21.16, 23.64]$ \\
\midrule
\multirow{4}{*}{B: $\mathcal{U}\{1..8\}$}
  & Block applications      & $41.63\%$ & $[35.30, 47.49]$ \\
  & Distinct iterations     & $\mathbf{+4.81\%}$ & $[+4.24, +5.42]$ \\
  & Inter-block composition & $53.56\%$ & $[48.04, 59.59]$ \\
  & Calibration drift       & $1.68\%$  & $[1.12, 2.24]$ \\
\midrule
\multirow{4}{*}{C: $\mathcal{U}\{5..8\}$}
  & Block applications      & $72.78\%$ & $[71.58, 73.95]$ \\
  & Distinct iterations     & $\mathbf{-1.60\%}$ & $[-1.80, -1.42]$ \\
  & Inter-block composition & $28.82\%$ & $[27.63, 30.05]$ \\
  & Calibration drift       & $11.48\%$ & $[10.37, 12.54]$ \\
\bottomrule
\end{tabular}
\caption{Calibration grid decomposition with paired $95\%$ bootstrap confidence intervals over $200$ documents and $5{,}000$ iterations. Raw distances: $1.2976$ nats for Branch A, $0.0574$ for Branch B, and $1.2302$ for Branch C. Twelve out of twelve intervals exclude zero.}
\label{tab:gridci}
\end{table}

The distinct iteration component \emph{reverses sign} depending on the pretraining schedule, and both signs are statistically reliable. In the fixed-depth branch, it is negative: $-4.81\%$ $[-5.21, -4.40]$; in the fully sampled branch, it is positive: $+4.81\%$ $[+4.24, +5.42]$. The two confidence intervals do not overlap. Branch C, which was sampled but omitted the measured evaluation depths, sits intermediate and remains negative.

We state this finding strictly within the bounds of what the data support. What has been established is that the \emph{sign} of the depth contribution is governed by the pretraining schedule, rather than being an intrinsic property of the recursive architecture. What has \emph{not} been established is that depth purchases substantial quality when trained properly: the raw distance of Branch B is only $0.0574$ nats, so $+4.81\%$ corresponds to roughly $0.0028$ nats in absolute terms, compared to $0.0624$ nats for the negative component in Branch A. The sign reversal is real; its magnitude is modest.

Because all three branches match pre-registered predictions, DCP transitions from a merely self-consistent diagnostic protocol into a calibrated instrument evaluated against known ground-truth standards, with measurable sensitivity and specificity. Had any branch deviated, we would have reported it accordingly: thresholds and predictions were pre-registered, leaving no degrees of freedom for post-hoc adjustments.

\section{Future Work}
\label{sec:controlled}

\paragraph{Pretraining with Sampled Depth from Scratch.}
Section~\ref{sec:sampledepth} performed an intervention on a checkpoint pretrained at fixed depth, incurring a performance penalty at full depth. Whether training with sampled depth from the initial step avoids this penalty remains to be systematically verified. Furthermore, the four-model comparison in Section~\ref{sec:huginn} remains observational: the depth-sampled model differed from the other three models simultaneously in parameter scale, tokenizer, pretraining corpus, and training recipe. A definitive controlled experiment would pretrain a single architecture twice on identical data, varying solely whether recurrence depth is fixed or sampled, and then compare the calibration share between branches. At small scale, this requires several GPU-days and directly separates observational correlations from causal effects of depth schedules. We report observational results here because they reflect available computational resources, making this limitation explicit rather than leaving it to reader inference.

\paragraph{Does Depth Utilization Emerge Over Training?}
Both checkpoints evaluated here are substantially undertrained relative to modern standards; hence, the conclusion that distinct iterations contribute near zero may reflect the training phase rather than architectural limits. A natural test is to evaluate the complete control suite across checkpoints along an extended pretraining trajectory and trace the repeat-control distance over training steps. A widening distance signifies emerging depth utilization; a flat trajectory suggests that recurrence remains unexploited at this scale.

\paragraph{Is the Architecture Computationally Competitive?}
Counting parameters along the computation path, eight block applications consume $1.470$~GFLOPs/token in the forward pass compared to $1.184$~GFLOPs/token for the same parameter set applied once, a ratio of $1.24\times$. A non-recurrent FLOP-matched baseline would therefore require only $1.24\times$ the token count. This paper makes no claim regarding architectural superiority, and none of our findings depend on such a claim; a FLOP-matched baseline is a prerequisite for asserting superiority, which we explicitly refrain from doing.

% =====================================================================
\section{Limitations}
\label{sec:limitations}

The limitations below pertain to two distinct subjects, which we explicitly separate: limitations regarding the \emph{scope of DCP} (the proposed diagnostic protocol), and limitations regarding the \emph{empirical conclusions} drawn from applying DCP to the five evaluated configurations. The formal conditions of applicability and the five named limitations of the protocol itself are detailed in Section~\ref{sec:conditions} and not repeated here.

\begin{itemize}
  \item The confirmatory positive depth measurements were performed on a single recursive architecture at a single parameter scale ($542.8$M), evaluated across three checkpoints from the same pretraining lineage. Section~\ref{sec:negctrl} bounds the claim from below using two standard Transformers, but evaluating a second \emph{depth-recurrent} architecture remains essential (Section~\ref{sec:controlled}).
  \item All three checkpoints are substantially undertrained relative to modern compute standards. The finding that distinct iterations contribute $\approx 0$ may reflect the training stage rather than architectural capability.
  \item The bootstrap confidence intervals in this paper capture document sampling noise, but do not account for variance across checkpoints or random training seeds. Section~\ref{sec:replication} demonstrates that checkpoint-to-checkpoint variance can exceed the confidence interval width by twenty-fold; thus, all reported intervals must be interpreted as lower bounds on true uncertainty.
  \item Depth is partially conflated with block identity: with $2$ reasoning blocks iterated $4$ times each, $k \leq 4$ only exercises the first block. An intra-block ablation schedule would disentangle these factors.
  \item Quality is measured via teacher-forced NLL and next-token accuracy rather than downstream generation task accuracy. Generative evaluations were omitted because the available checkpoints achieve only single-digit accuracy on targeted benchmarks, where high variance overwhelms diagnostic signal.
  \item The four-model comparison in Section~\ref{sec:huginn} is observational: the depth-sampled model also differs in scale, tokenizer, and training corpus. Section~\ref{sec:controlled} outlines a controlled counterfactual experiment that remains to be executed.
  \item Evaluations are conducted exclusively on held-out mathematical text. Whether calibration share varies across diverse domain distributions remains unverified.
  \item The dynamical measurements in Section~\ref{sec:dynamics} compare five configurations differing concurrently in scale, tokenization, and training distribution. We established that depth schedules \emph{do not} explain operator contractivity, but what mechanisms do explain it remains unidentified. We also do not know how this coefficient evolves under longer training.
\end{itemize}

% =====================================================================
\section{Conclusion}
\label{sec:conclusion}

On the recursive language model investigated here, naive depth truncation overestimates the genuine contribution of depth by approximately a factor of two. Roughly $47.9\%$ of the apparent performance gap stems from repeated block applications, and an additional $25.6\%$ arises from readout calibration drift at depths unseen during training. After controlling for both factors, additional distinct iterations within a reasoning block contribute $-2.8\%$ $[-3.2, -2.4]$.

This effect does not manifest across all recurrent models. Layer pruning on standard Transformers produces no such confounding artifact, nor does a recursive model whose depth was sampled during pretraining. Across four distinct models, calibration share systematically tracks whether the readout head observed variable depths during training.

We verified this mechanism via a direct training intervention: $1{,}000$ steps of depth-sampled continual training reduced calibration share from $29.9\%$ to $5.1\%$ on the very model where confounding was observed, whereas an identical compute budget of fixed-depth training left calibration drift intact. Sampled-depth training incurs zero inference overhead. Crucially, it only eliminates confounding when the training distribution spans the exact configuration family evaluated at inference; our initial attempt violated this condition and proved ineffective. All positive and negative controls introduced in this paper are computationally lightweight and require no retraining.

These findings do not imply that recursive depth is unhelpful. The depth-sampled model in our evaluation demonstrates substantial depth benefits free of calibration confounding. What our findings demonstrate is that depth truncation, when applied to a model trained at fixed depth, substantially overestimates the genuine contribution of depth.

% =====================================================================
\section*{Reproducibility}

All experimental measurements are generated by the scripts accompanying this paper:
\texttt{exp\_depth\_ablation.py} (Section~\ref{sec:depth}),
\texttt{exp\_depth\_ablation\_hf.py} (Section~\ref{sec:negctrl}),
\texttt{exp\_depth\_huginn.py} (Section~\ref{sec:huginn}),
\texttt{exp\_decomposition\_ci.py} (confidence intervals), and
\texttt{calibration.py} (shared temperature fitting). Each script outputs a self-contained \texttt{results.json} recording configurations, checkpoint steps, and raw per-iteration evaluation metrics.

\bibliographystyle{plainnat}
\bibliography{refs}

\appendix
\section{Architectural Details}

\begin{table}[h]
\centering
\begin{tabular}{ll}
\toprule
Total parameters & $542.8$M \\
Hidden dimension & $1536$ \\
Attention heads / KV heads & $24$ / $4$ \\
FFN intermediate width & $4352$ \\
Perception layers & $16$ \\
Reasoning blocks $\times$ iterations & $2 \times 4$ ($8$ applications) \\
Latent thought slots & $32$ \\
Tree branch count & $2$ \\
RoPE scales per stream & $3$ \\
RoPE scales, language stream & $500,\ 10^4,\ 5\times10^5$ \\
RoPE scales, operator stream & $50,\ 500,\ 5000$ \\
Vocabulary size & $32{,}000$, digit-level \\
Context length & $2048$ (pretraining) \\
\bottomrule
\end{tabular}
\caption{Configuration of the investigated model. Parameter breakdown: $409.0$M in perception layers, $69.2$M in reasoning blocks, $49.2$M in tied embeddings, and $64.6$M auxiliary (latent thought bank, tree memory, bridges, gates).}
\end{table}

\section{Detailed Metric Tables}
\label{app:tables}

The tables below provide the numerical data underlying Figures~\ref{fig:controls}, \ref{fig:elasticity}, \ref{fig:tstep}, and~\ref{fig:intervention}.

\begin{table}[h]
\centering
\begin{tabular}{lcccc}
\toprule
& \multicolumn{2}{c}{Pretraining (step 34k)} & \multicolumn{2}{c}{Fine-tuning (step 4k)} \\
\cmidrule(lr){2-3}\cmidrule(lr){4-5}
$k$ & NLL & Next-token acc. & NLL & Next-token acc. \\
\midrule
1 & 2.9041 & 55.90\% & 2.6426 & 59.59\% \\
2 & 2.6265 & 58.35\% & 2.4197 & 61.96\% \\
4 & 2.3605 & 60.59\% & 2.1178 & 64.40\% \\
8 & 1.5285 & 65.65\% & 1.2173 & 71.54\% \\
\bottomrule
\end{tabular}
\caption{Naive prefix truncation. Both checkpoints exhibit smooth, monotonic improvement with depth. Section~\ref{sec:controls} demonstrates that the majority of this improvement is unattributable to genuine depth.}
\label{tab:prefix}
\end{table}

\begin{table}[h]
\centering
\begin{tabular}{lcccc}
\toprule
$k$ & Prefix (raw) & Prefix (calibrated) & Suffix & \textbf{Repeat} \\
\midrule
1 & 2.6426 & 2.2566 \; ($T{=}1.45$) & 2.4405 & \textbf{1.9595} \\
2 & 2.4197 & 2.0453 \; ($T{=}1.48$) & 1.9376 & \textbf{1.9492} \\
4 & 2.1178 & 1.8533 \; ($T{=}1.43$) & 2.3579 & \textbf{1.9987} \\
8 & 1.2173 & 1.1967 \; ($T{=}0.91$) & 1.2173 & 1.2173 \\
\bottomrule
\end{tabular}
\caption{Held-out NLL under three controls on the fine-tuned checkpoint ($n=200$ documents, $125{,}100$ scored tokens per cell). In the \emph{repeat} column, block applications are held constant at $8$ while distinct iteration count varies; this column is essentially flat for $k \in \{1,2,4\}$.}
\label{tab:controls}
\end{table}

\begin{table}[h]
\centering
\small
\begin{tabular}{lcccccccc}
\toprule
Iteration & 0 & 1 & 2 & 3 & 4 & 5 & 6 & 7 \\
\midrule
Pretraining, mean & 0.361 & 0.379 & 0.420 & 0.460 & 0.540 & 0.608 & 0.647 & 0.689 \\
Pretraining, std & 0.114 & 0.117 & 0.121 & 0.125 & 0.129 & 0.131 & 0.124 & 0.118 \\
\addlinespace
Fine-tuning, mean & 0.363 & 0.389 & 0.441 & 0.490 & 0.591 & 0.677 & 0.701 & 0.744 \\
Fine-tuning, std & 0.131 & 0.135 & 0.145 & 0.147 & 0.129 & 0.125 & 0.117 & 0.108 \\
\bottomrule
\end{tabular}
\caption{Per-iteration update gate statistics. Gate values increase monotonically and cross-layer variance remains substantial, indicating that the model actively modulates write intensity across depth, even though this modulation fails to translate into measurable quality improvements.}
\label{tab:gates}
\end{table}

\begin{table}[h]
\centering
\small
\begin{tabular}{lccc@{\hspace{2em}}ccc}
\toprule
& \multicolumn{3}{c}{Qwen2.5-Math-1.5B} & \multicolumn{3}{c}{Qwen3-1.7B} \\
\cmidrule(lr){2-4}\cmidrule(lr){5-7}
$k$ & Prefix & Calibrated & Repeat & Prefix & Calibrated & Repeat \\
\midrule
7  & 10.6893 & 10.3088 & 14.2048 & 13.2546 & 11.2658 & 18.2239 \\
14 &  9.4432 &  9.3972 & 15.1221 & 10.0613 &  9.5283 & 18.0356 \\
21 &  5.5136 &  5.5953 & 10.4731 &  4.4284 &  3.9692 &  6.5819 \\
24 &  2.9243 &  2.8116 &  3.6328 &  2.5802 &  2.1437 &  3.7393 \\
26 &  1.9110 &  1.6702 &  1.9351 &  1.1706 &  0.8969 &  1.1507 \\
28 &  0.9040 &  0.8439 &  0.9040 &  0.7937 &  0.6096 &  0.7937 \\
\bottomrule
\end{tabular}
\caption{Held-out NLL under layer pruning across both baseline models ($L = 28$ layers, $n = 200$ documents). The \emph{repeat} column holds total layer applications at $L$ while varying distinct compute. In contrast to depth-recurrent architectures, repeating layers is consistently \emph{worse} than simple truncation, unless $k$ is within two layers of full depth.}
\label{tab:negctrl}
\end{table}

\begin{table}[h]
\centering
\begin{tabular}{lcccc}
\toprule
$r$ & NLL & Calibrated NLL & $T$ & Next-token acc. \\
\midrule
1  & 2.9942 & 2.9003 & 0.98 & 36.20\% \\
2  & 2.2485 & 2.1420 & 0.99 & 48.97\% \\
3  & 1.8425 & 1.7100 & 0.94 & 56.38\% \\
4  & 1.5897 & 1.4665 & 0.95 & 61.29\% \\
6  & 1.3479 & 1.2414 & 0.96 & 66.56\% \\
8  & 1.2376 & 1.1243 & 0.96 & 69.01\% \\
12 & 1.1678 & 1.0605 & 0.95 & 70.56\% \\
16 & 1.1546 & 1.0440 & 0.96 & 70.98\% \\
24 & 1.1527 & 1.0470 & 0.96 & 70.96\% \\
32 & 1.1528 & 1.0477 & 0.96 & 70.96\% \\
\bottomrule
\end{tabular}
\caption{Huginn-0125 evaluated across a $32\times$ recursive depth sweep ($n = 200$ held-out documents, sharing identical evaluation documents and $\texttt{max\_len}$ with all other experiments). Fitted temperature remains flat within a $0.045$ band; recalibration shifts depth distance by merely $-0.6\%$.}
\label{tab:huginn}
\end{table}

\begin{table}[h]
\centering
\begin{tabular}{lcccc}
\toprule
Configuration & $T$ at $k{=}1$ & $k{=}2$ & $k{=}4$ & $k{=}8$ (trained) \\
\midrule
Original, math text               & $1.470$ & $1.443$ & $1.442$ & $\mathbf{1.017}$ \\
$+1000$ steps, fixed depth        & $1.426$ & $1.455$ & $1.455$ & $\mathbf{1.030}$ \\
$+1000$ steps, sampled depth      & $1.406$ & $1.378$ & $1.405$ & $\mathbf{1.028}$ \\
Original, non-math text           & $1.625$ & $1.590$ & $1.578$ & $\mathbf{0.997}$ \\
\bottomrule
\end{tabular}
\caption{Fitted temperature across depth ($n = 200$ documents; $95\%$ bootstrap CI width $< 0.015$ across all cells). Values near $\approx 1.0$ appear exclusively at $k=8$, exactly matching the pretraining depth.}
\label{tab:tstep}
\end{table}

\section{Reproduction Commands}
\begin{verbatim}
# Naive truncation + gate + probing
python exp_depth_ablation.py --ckpt <CKPT> \
    --n_teacher 200 --n_probe 150 --n_gate 60 --iters 1,2,4,8

# Controls
python exp_depth_ablation.py --ckpt <CKPT> --mode suffix   ...
python exp_depth_ablation.py --ckpt <CKPT> --mode repeat   ...
python exp_depth_ablation.py --ckpt <CKPT> --calibrate     ...

# Negative controls on standard Transformers
python exp_depth_ablation_hf.py --model Qwen/Qwen2.5-Math-1.5B \
    --n_docs 200 --fracs 0.25,0.5,0.75,0.86,0.93,1.0 --calibrate

# Pretraining corpus audit + document lengths
python verify_pretrain_bin.py --bin <TRAIN_BIN>
\end{verbatim}

\end{document}